\documentclass[11pt]{article}
\usepackage{acl}

\usepackage{times}
\usepackage{latexsym}

\usepackage[utf8]{inputenc}
\usepackage[T1]{fontenc}
\usepackage{amsmath, amssymb}
\usepackage{graphicx}
\usepackage{booktabs}
\usepackage{microtype}
\usepackage{algorithm}
\usepackage{algpseudocode}
\usepackage{subfigure}
\usepackage{float}
\usepackage{placeins}

\usepackage[scaled=0.85]{beramono}
\usepackage[T1]{fontenc}
\usepackage{url}

\title{Factual Retrieval in LLMs \\ Is a Redundant, Distributed and Non-Contiguous Process}

\author{
  Hail Hochman \\
  Bar-Ilan University \\
  \texttt{hailzanbar1@gmail.com}
  \And
  Natalie Shapira \\
  Northeastern University \\
  \texttt{nd1234@gmail.com}
  \And
  Yoav Goldberg \\
  Bar-Ilan University \\
  \texttt{yoav.goldberg@gmail.com}
}

\begin{document}
\maketitle

\begin{abstract}
Large language models (LLMs) store and recall factual knowledge, yet the precise mechanism of how entity representations are transformed to enable specific attribute retrieval remains underexplored. In this work, we investigate this mechanism through the lens of an ``attribute-computation path''—a sequence of computational steps over the entity representation required to elicit a target attribute. We then propose an iterative patching protocol to identify a minimal subset of layers necessary for this computation. Applying our method to LLaMA 3.1 8B and Qwen3 8B, we find that these paths are non-contiguous, often skipping layers, and that models possess multiple, functionally-equivalent paths for the same entity and fact, highlighting a high degree of redundancy in attribute computation. This implies that knowledge computation is highly distributed, potentially explaining the localization-editing mismatch and suggesting that knowledge storage and retrieval in LLMs is far from being well understood.
\end{abstract}

\section{Introduction}
\label{sec:intro}

Large language models (LLMs) have been shown to store and recall factual knowledge expressed as entities and their relations \citep{petroni2019language, jiang2020can, cohen2023crawling}. For example, when prompted with \textit{``The mother tongue of Angela Merkel is''}, an LLM will predict \textit{``German''}, the correct answer. While prior studies attempted to identify components of the model where factual knowledge is stored \citep{geva2021transformer, dai2021knowledge, geva2022transformer,meng2022locating, gurnee2023language, katz2024backward, yu2024neuron} and to map the general information flow during recall \citep{geva2023dissecting, nanda2023factfinding, chughtai2024summing, wang2025functional, yao2024knowledge, yu2025back}, the precise mechanics of how facts are retrieved from model parameters remain unclear. Existing work posits that this process occurs at the last entity token position; however, the term ``retrieval'' itself is not well defined. It remains ambiguous whether retrieval is a continuous process of information accumulation across a range of layers, or if it is defined by a discrete state---a specific point where the representation becomes rich enough to elicit the specific correct attribute. Furthermore, the transformations the entity representation undergoes \textit{before} it becomes capable of eliciting the target attribute are largely unmapped. To address these complexities, we propose investigating the factual recall mechanism through a new lens: mapping the \emph{minimal computation path} the entity representation must undergo in order to achieve attribute recall, in a given factual-recall prompt.

\begin{figure}[t]
    \centering
    \includegraphics[width=1\linewidth]{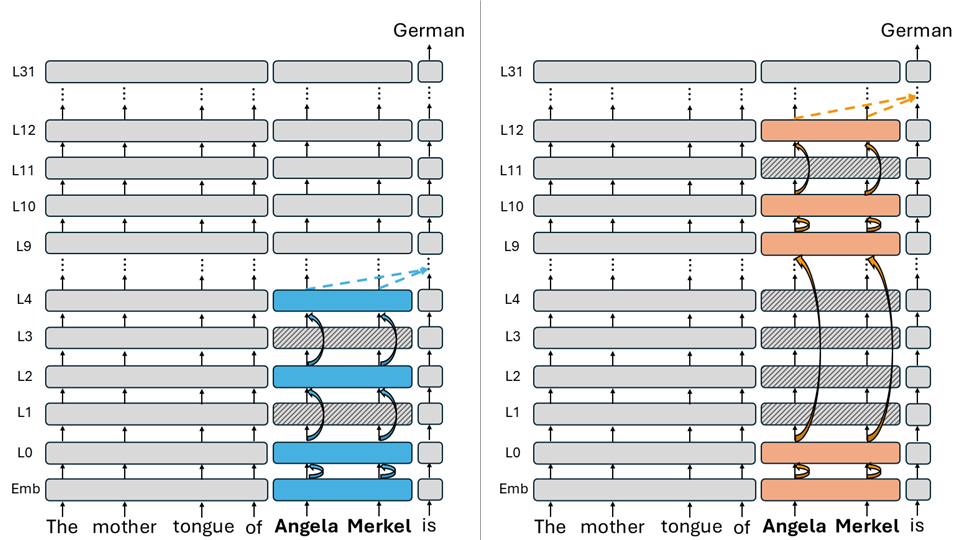}
    \caption{Two functionally equivalent minimal attribute-computation paths for the entity ``Angela Merkel''. Both paths are non-contiguous, skipping intermediate layers. The blue path represents a compact, early computation, while the orange path illustrates a deeper, more distributed alternative.
    \vspace{-6pt}}
    \label{fig:redundant_paths}
\end{figure}

Given a prompt with an entity and a relation (e.g., \textit{``The mother tongue of Angela Merkel is''}), and a target attribute (e.g., \textit{``German''}), we define an \textit{attribute-computation path} as a sequence of computations over the entity representation, that together enable the LLM to answer the prompt correctly. Each computation is implemented as a transformer layer, and we seek a subset of layers which is both sufficient and necessary for computing the desired output. We call such a subset of layers, in which no layer can be removed without hurting the computation, \emph{a minimal attribute computation path}, and observe that each layer along such a minimal path plays an active role in the factual recall process for the attribute.

We present a method to identify minimal computation paths for a given factual recall prompt, and use it to extract computation paths for a diverse set of such prompts, in two open-weight LLMs (LLaMA 3.1 8B and Qwen3 8B). Our experiments reveal that for a  majority of target attributes:
\begin{itemize}
\setlength\itemsep{0.1em}
\item \emph{The minimal path naturally ends in the early-mid layers}, after which the attribute value is fully available and no further processing is required. This is consistent with observations in previous works \cite{geva2023dissecting, nanda2023factfinding, meng2022locating}.
\item \emph{Minimal paths consist of more than a single layer}: there isn't a single layer responsible for knowledge retrieval for a given entity-relation-attribute triplet, but rather the entity must be processed by multiple layers before the information is fully available.
\item \emph{Minimal paths are often sparse and non-contiguous}: only a subset of layers participate in each factual recall process, while others may be skipped (but the same layer may participate in multiple factual recall processes).
\end{itemize}
Taken together, the last two items indicate that factual knowledge is stored in a distributed and non-localized manner: there isn't a single layer or range of layers we can say is responsible for a given fact, rather the factual information is distributed across the parameters of multiple layers.
Our experiments further reveal ``backup'' paths \citep{wang2022interpretability, mcgrath2023hydra}: 
\begin{itemize}
\item \emph{Minimal paths are not unique}: for most facts we explored, there is at least one alternative set of layers that is sufficient for retrieving the factual information. These alternate paths are deeper and longer than the primary paths the model uses in natural, non-intervened runs (Figure \ref{fig:redundant_paths}).
\end{itemize}
We find that the same factual knowledge can be retrieved through multiple distinct computational paths within the model. This suggests an even more complex and elaborate form of knowledge representation, where the storage and retrieval mechanism is not only distributed but also highly redundant. 
These findings provide a new perspective on the mechanism of factual recall \citep{geva2023dissecting, meng2022locating, yu2025back}.
We detail these, together with additional findings, in Section \ref{sec:experiments}.
To locate the minimal paths, we use a novel iterative activation-patching protocol, described in Section \ref{sec:method}.

In summary, our work makes three main contributions.
\begin{itemize}
    \setlength\itemsep{0.1em}
    \item We offer a new perspective on the sub-process of entity enrichment \cite{geva2023dissecting,yu2025back}, showing that the computation of the entity representation for specific attribute recall unfolds as a \emph{multi-step process} involving several necessary intermediate transformations, while at the same time not requiring most model layers.
    \item We reveal that \emph{multiple minimal and functionally equivalent attribute computation paths exist} within the model with different lengths and in different regions, highlighting the distributed and redundant nature of factual knowledge representation.
    \item We propose a novel activation patching protocol to identify a minimal subset of layers required for computation of an attribute over the entity activations, providing a precise view of how factual knowledge enriches the entity representations. 
\end{itemize}

Based on our findings, we propose an explanation for the discrepancy between where the facts are located and where they are most effectively edited \citep{hase2023does}. We suggest that because attribute computation requires multiple transformations of the entity representation, editing succeeds by intervening at a necessary transformation stage, not necessarily where the knowledge is localized.
Our code is available at \url{https://github.com/hhochman/llm-factual-retrieval}.

\section{Related Work}

\paragraph{Entity Representation \& Factual Recall.}
Prior work frame factual recall in large language model as a three-stage process \citep{geva2023dissecting, yu2025back, chughtai2024summing, nanda2023factfinding}. In their framing, early attention layers first \emph{consolidate} the representation of the entities initial tokens into its final token \citep{nanda2023factfinding}. Then, the final entity token undergoes an \emph{entity enrichment phase} in which various entity attributes are loaded from the model parameters and encoded on the (final) entity tokens. This stage is primarily driven by early feed-forward (MLP) sublayers \citep{meng2022locating, geva2023dissecting, yu2025back} in which the knowledge resides. Finally, once entity enrichment concludes and the knowledge is retrieved and loaded into the entity representation, information from the relation tokens propagates forward in the sequence, reaching the final token of the prompt. 
Then, the representation at the final token of the prompt \textit{queries} the enriched entity representation to extract the target attribute \citep{geva2023dissecting}.

It is also established that the resulting representation at the entity's final token plays a central role in knowledge storage and retrieval \citep{geva2023dissecting, yu2025back, hernandez2023linearity, ghandeharioun2024patchscopes}.
Recent work demonstrates that, for many relations, attributes can be decoded from the entity’s final token using a simple, approximately linear transformation \citep{hernandez2023linearity}.
Furthermore, probing the hidden states of this token in late--mid layers can reveal the extent of the model's stored knowledge about the entity, without relying on generated outputs \citep{gottesman2024estimating}. When the relation follows the entity in the prompt, enrichment and attribute extraction can occur not only at the entity position but also at the relation and final positions \citep{yu2025back, chughtai2024summing}. 
In these cases, deeper attention and feed-forward components store factual knowledge linking the entity and relation to the attribute.

While \citealp{meng2022locating} identified stages where representations become sufficient for attribute restoration and hypothesized accumulation across MLPs, the precise causal computations of this process remain under-explored.
We zoom in on the entity enrichment stage and trace the complete causal paths by which individual facts are loaded into the entity representation. Our findings reveal a complex and elaborate system in which each individual fact requires processing by multiple LLM layers before it can be accessed: there is no single layer in which a specific fact is retrieved.

\paragraph{Circuit Redundancy.}
Early work in mechanistic interpretability of LLMs and other models focused on identifying circuits: subgraphs of the model's computational graph responsible for specific behaviors \citep{olah2020zoom, elhage2021mathematical, wang2022interpretability, goldowsky2023localizing, ameisen2025circuit}.
However, the definition of a ``circuit'' is complicated by the distributed and redundant nature of LLMs. 

\citet{mcgrath2023hydra} revealed the ``hydra effects'', where the model self-repairs by activating redundant components when primary ones are ablated. \citet{wang2022interpretability} similarly identified ``backup heads'' that perform the same function as primary heads but are only active when the primary path is disrupted. This complexity extends to model editing; \citet{hase2023does} demonstrated that while facts can be successfully edited at specific model locations, the most effective editing targets often diverge from the locations identified by standard localization methods. These findings suggest that models do not rely on a single, unique circuit for a given task, but rather possess multiple functionally equivalent mechanisms.

While \citet{wang2022interpretability} revealed backup ``Name-Mover Heads'' in the IOI task, and the hydra effect \citep{mcgrath2023hydra} has been investigated primarily regarding the information flow to the final token, we demonstrate redundancy within a different factual recall mechanism: the layers performing the computation process over the entity tokens that is used to retrieve the attribute.

\section{Preliminaries and Notations}
\label{sec:preliminaries}

\paragraph{Layers and Activations.}
Let $I^{(i,\ell)}$ denote the input activations to layer $\ell$ at position $i$.\footnote{With slight abuse of notation, we use a single index to refer to either a single token position or a range of tokens (corresponding to an entity).} A layer function $L_\ell$ applied to $I^{(i,\ell)}$ results in the layer's output $O^{(i,\ell)}$, which serves as the input to the next layer, $I^{(i,\ell+1)}$:
\begin{align*}
L_\ell(I^{(i,\ell)}) = O^{(i,\ell)} = I^{(i,\ell+1)}
\end{align*}
We refer to the same activations as either $O^{(i,\ell)}$ or $I^{(i,\ell+1)}$, depending on the context.
The layer function $L_\ell$ computes attention ($Attn$) and MLP with residual connections:
\begin{align*}
L_\ell(I^{(i,\ell)}) &= I^{(i,\ell)} + a_{(\le i,\ell)} + m_{(i,\ell)} \\
a_{(\le i,\ell)} &= Attn_{\ell}(I^{(1,\ell)},...,I^{(i,\ell)}) \\
m_{(i,\ell)} &= MLP_{\ell}(I^{(i,\ell)} + a_{(\le i,\ell)})
\end{align*}

\paragraph{Runs and Activation Patching.}
We denote the prediction of a model M given prompt P as M(P). In this work we restrict ourselves to single word predictions using greedy decoding.

Given an $n$ tokens input prompt, the transformer decoder step with $|L|$ layers computes the values $O^{(i,\ell)}$ for $1 \le \ell \le |L|; 1 \le i \le n$, in topological order where a node $O^{(i,\ell)}$ is computed after nodes with positions $\le i$ and layers $< \ell$. 
A \emph{patch operation} $\Phi$ specifies a list of locations $(i,\ell)$ and a corresponding patch vector for each. A \emph{patched run} $M(P \mid \Phi)$ on an input prompt P computes the activation vectors according to their topological order, but, whenever $(i,\ell)$ corresponds to a patching location, $I^{(i,\ell)}$ is replaced with the corresponding value instead of its computed value, and the computation continues.

\section{Setup and Goal}

To understand how factual knowledge is stored and retrieved in LLMs, we are interested in tracing the computations performed by a transformer-based LLM when completing \emph{factual recall prompts} such as \textit{``The mother tongue of Angela Merkel is''}. We focus our attention on the process in which the \emph{entity representation} evolves through the layers until it is sufficient for the model to produce the correct answer (``German''). Considering each layer as performing a computation over the entity representation, we look for what we call a \emph{minimal attribute computation path}: a subset of layers that are \emph{required} for producing the correct attributes.

\paragraph{Factual Recall Prompts.}
We study prompts of the form $(e, r) \to a$, where $e$ is a subject entity, $r$ is a relation, and $a$ is the target attribute to be predicted (not part of the prompt). For example, in the above prompt
the subject entity $e$ is ``Angela Merkel'', the relation $r$ is ``mother tongue'', and the target attribute $a$ is ``German''. The relation may appear either before the entity or after it. Both entities and relations are drawn from a variety of domains. 

\paragraph{Entity Representation} is the activation values at the entity position for a given prompt. It evolves through the layers, with the representation at layer $\ell$ denoted as $O^{(e,\ell)}$.

\paragraph{Computation Paths.} 
A \emph{computation path} is a series of computations (a subset of layers) applied to an entity representation.  An \emph{attribute computation path} is a computation path after which the model correctly predicts the target attribute.
A \emph{minimal attribute-sufficient computation path} (or \emph{minimal attribute path} for short) is a path in which every included layer is essential; removing any layer from this set disrupts the computation of the attribute over the entity activations.\footnote{Note that "minimal" refers to the irreducability of the path rather than its length: multiple distinct minimal paths of different lengths may exist. Indeed, we show that this is often the case in practice.}

\paragraph{Relations to Knowledge Retrieval.}
Each layer belonging to a minimal attribute path plays an essential role in the process of \emph{knowledge retrieval} for that attribute: it either extracts (parts of) the attribute value from the model's parameters; transforms the entity representation to facilitate the knowledge extraction; or transforms extracted values into a form from which the attribute can be decoded.
Thus, we consider the study of minimal attribute paths as an essential milestone for understanding factual knowledge retrieval in transformer-based LLMs.

\begin{figure*}[t]
    \vspace{-4mm}
    \centering
    \includegraphics[width=1\textwidth]{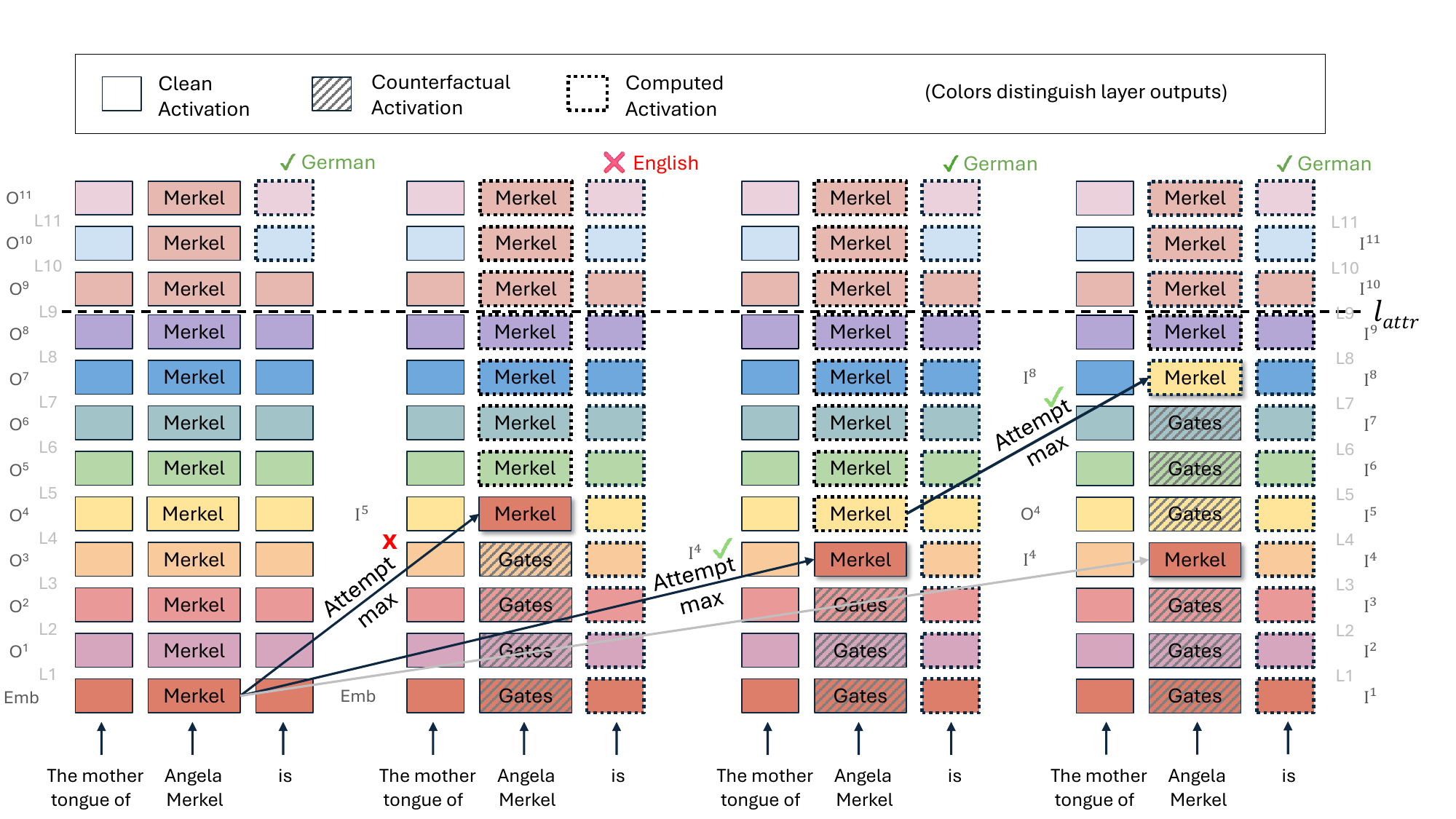}
        \caption{\textbf{Iterative Greedy Search for Minimal Computation Paths.} 
            We illustrate the process using the original prompt \textit{``The mother tongue of Angela Merkel is''} (Target: \textit{``German''}) and the counterfactual prompt \textit{``The mother tongue of Bill Gates is''} (Target: \textit{``English''}).
            First, $L_9$ is established as the attribute sufficient layer using the \emph{lock} operation. Then,
            \textbf{Failed Attempt (Left Arrow):} The algorithm attempts a maximal jump from the embeddings to layer 5, but the model predicts the counterfactual target, marking this skip as invalid. \textbf{Successful Attempt (Middle Arrow):} The algorithm identifies layer 4 as the highest layer into which a jump recovers the correct target. \textbf{Next Iteration (Right Arrow):} With layer 4 established as a necessary step, treating its output as the new source, the search repeats to identify the next maximal jump (to layer 8).
    \label{fig:primary_path_visualization}}
\end{figure*}

\section{Method}
\label{sec:method}

In an $|L|$-layers transformer $M$ for which $M(P_{(e,r)}) = a$,
the path $[L_1,...,L_{|L|}]$ is attribute sufficient for $a$. But are all the layers necessary? 

To identify a minimal attribute path, we start by determining the termination point of the path: the earliest layer $\ell_{attr} \le |L|$ after which no more computations on the entity representation are needed for $M$ to produce the attribute $a$ at the last token (\autoref{subsec:rep-ident}). Second, using layer $\ell_{attr}$ as an upper bound, we iteratively identify a minimal sequence of layers that progressively transform the entity embedding for recalling attribute $a$ (\autoref{subsec:path-ident}). Finally, we show that by replacing the upper bound $\ell_{attr}$ with $|L|$ and using the same iterative procedure, we can identify alternative minimal attribute paths (\autoref{subsec:extended-path}).

\subsection{Locating $\ell_{attr}$}
\label{subsec:rep-ident}

To bound the minimal attribute path from above, we seek a layer $\ell$ whose output does not need to be transformed by subsequent layers to produce attribute $a$. Removing a layer from the end of the path amounts to overriding its output with that of the previous layer. This gives rise to an operation we call \emph{lock}: 
\[lock(O^{(e, {\ell})}) :=  O^{(e, {\ell+k})} \underset{patch}{\leftarrow} O^{(e, \ell)} \quad \forall k > 0 \]
This is an instance of activation patching  \citep{vig2020investigating, wang2022interpretability, meng2022locating} which we call \emph{activation locking}: patching the activations from an entity representation at layer $\ell$ to all higher layers of the same entity.

For a given prompt $P$, we seek the earliest layer on which we can apply lock operation and retain the same correct attribute $a$:
\[\ell_{attr} = \min_{\ell} \, \text{s.t.} \, M\bigl(P \mid lock(O^{(e, {\ell})})\bigr) = M(P) \]

\noindent Fig \ref{fig:l_attr_visualization} (Appendix~\ref{app:visualizations}) illustrates this process.

\subsection{Minimal Path Identification}
\label{subsec:path-ident}

Having identified $\ell_{attr}$, we seek to identify a path (a sequence of layers) $[L_{\alpha_1}, \ldots, L_{\alpha_k}]$, $\alpha_i < \alpha_{i+1}; \; \alpha_k \le \ell_{attr}$, which forms a subset of the model’s layers constrained by the upper bound $\ell_{attr}$. This sequence represents a minimal set of layers such that, when the entity embedding is processed only through these layers, the model $M$ still produces the desired output $a$. 

To identify non-essential layers, we use counterfactual prompts: prompts of the form $(\hat{e},r) \to \hat{a}$ where $\hat{e}$ shares the same semantic type, number of tokens and sequence positions with $e$, but for which $M$ produces an answer $\hat{a} \neq a$.  
We propose a heuristic: if a layer can process counterfactual entity data without disrupting the final attribute prediction, this layer is not necessary for the computation. Thus, given counterfactual entity activations $O_{\text{counter}}^{(e,\ell)}$ from running $M$ on the counterfactual prompt, and a candidate path, we define an activation patching operations called \emph{isolate} in which: (a) the first path layer $L_{\alpha_1}$ receives the clean entity embedding $E^{(e)}_{\text{clean}}$; (b) each subsequent path layer $L_{\alpha_i}$ receives the output of the previous path layer $L_{\alpha_{i - 1}}$; and (c) all layers between path layers receive the corresponding activations from the counterfactual run. Formally:

\vspace{5pt}
\noindent$isolate( layers = {[\alpha_1, ..., \alpha_k]}) := $
$$
:= \begin{cases}
I^{(e, \ell=\alpha_1)} \underset{patch}{\leftarrow} E^{(e)}_{\text{clean}} & \ell = \alpha_1 \\
I^{(e, \ell=\alpha_{i})} \underset{patch}{\leftarrow} O^{(e, \alpha_{i-1})}_{\text{computed}} & \ell = \alpha_i \in \text{layers} \\
I^{(e, \ell)} \quad\; \underset{patch}{\leftarrow} I^{(e, \ell)}_{\text{counter}} & \ell \not\in \text{layers}, \ell < \alpha_k\\
\end{cases}
$$
where $O_{\text{computed}}$ are the non-patched values computed in the current run. Steps (a) and (b) ensure that only the layers within the candidate path contribute to the attribute computation over the entity tokens. Meanwhile, step (c) ensures that any information propagating to subsequent positions from the bypassed layers is counterfactual, demonstrating that these intervening layers are not necessary for the factual information flow.
The rightmost column in Figure \ref{fig:primary_path_visualization} visualizes this operation for \emph{isolate([4,8])}: the path is isolated by patching the output of each path layer to the input of the next path layer (e.g., patching $E^{(e)}_{\text{clean}}$ into $I^{(e,4)}$ and $O^{(e,4)}_{\text{computed}}$ into $I^{(e,8)}$), while intermediate layers are patched with counterfactual values.

Given a prompt ${P}_{(e, r) \to a}$ and $\ell_{attr}$ we say that a path prefix $[\alpha_1, ..., \alpha_k]$, $\alpha_k \le \ell_{attr}$ is a \emph{valid} prefix of a \emph{computation path} iff:

\begin{multline*}
M\Bigl(P \mid isolate([\alpha_1, ..., \alpha_k])) = \\
\shoveleft{M\Bigl(P \mid isolate([\alpha_1, ..., \alpha_k]), lock(O^{(e, {\ell_{attr}})})\Bigr)} \\
= M(P) = a
\end{multline*}

We construct a minimal computational path via an iterative greedy search starting from the entity embedding\footnote{In our analysis (e.g., when calculating path length), we consider the entity embedding $E$ as the implicit start of all paths.}. If the embedding layer itself is sufficient (i.e., locking $E^{(e)}$ yields $a$), the path is empty. Otherwise, We start with an empty prefix (feeding the clean embedding through all the layers) and at each step, we extend the prefix by skipping as many layers as possible after its last layer while remaining a valid prefix. This strategy maximizes the number of skipped layers at every iteration, ensuring the final sequence is minimal. Formally, we define the extension operation:
\noindent\begin{multline*}
\text{extend}(\text{path}) := \text{path} \oplus \text{next}(\text{path}) \hfill \\
\quad \text{next}([\alpha_1,\ldots,\alpha_{k-1}]) := \hfill \\
\max_{\alpha_{k-1} < \alpha_k \le \ell_{attr}} \text{ s.t. } ([\alpha_1,\ldots,\alpha_{k-1}, \alpha_k] \text{ is valid})
\end{multline*}
and repeatedly apply \emph{extend}, starting from an empty path, until either the last path layer $\alpha_k = \ell_{attr}$ or 
$M(P | lock(O^{(e, {\alpha_k})})) = MP(P) = a$.\footnote{We allow early stopping when $\alpha_k < \ell_{attr}$ as skipping intermediate layers may allow the path to reach sufficiency earlier than the initial upper bound. In such cases, we redefine $\ell_{attr} = \alpha_k$ for later analyses.}
The process is illustrated in Figure \ref{fig:primary_path_visualization}.

\subsection{Alternative Minimal Path}
\label{subsec:extended-path}
The procedure in \autoref{subsec:path-ident} searches for a minimal path that reaches the sufficient representation in $\ell_{attr}$. What if we relax this restriction, and allow the search procedure to construct paths that end after layer $\ell_{attr}$? Under this definition a path prefix is valid if $M(P \mid isolate([\alpha_1, ..., \alpha_k])) = M(P) = a$ and we stop the prefix extension process when locking the last prefix layer yields the correct attribute.

This procedure reveals layers capable of executing the attribute computation even when the information integration occurs at a different depth than in the original forward pass.

\section{Experiments and Results}
\label{sec:experiments}
\noindent\textbf{Models.}
we experiment with two decoder-only transformer models: LLaMA 3.1 8B (32 layers) \citep{dubey2024llama} and Qwen3 8B (36 layers) \citep{yang2025qwen3}.

\noindent\textbf{Data}
We use the CounterFact dataset \citep{meng2022locating}, which provides prompts in the format described above, each paired with the correct target attribute. For each model, we select 2,000 prompts for which the target attribute is a single token and this token is correctly predicted by the model.

\subsection{Main Experiments}
\label{subsec:main_exp}

We applied our path identification method to the dataset to extract the causal computation paths for all entity--attribute pairs. We implemented the interventions using the \texttt{NNsight} package \citep{fiotto-kaufman2025nnsight}.

\paragraph{When is attribute knowledge available?} 
For both models, using the \emph{lock} operation, we find that the attribute computations complete in the early-to-mid layers, and further computation at the entity tokens is not required to produce the correct answer. The average $\ell_{attr}$ is 4.61 for LLaMA and 7.97 for Qwen, although for some cases we reach layer 15 and even 20 for both models (See Appendix \ref{app:quantitative_analysis} Figure \ref{fig:experiment1_results}(a) for the complete distribution). 
This aligns with previous studies on factual recall ~\citep{geva2023dissecting, yu2025back} that also place the conclusion of entity enrichment (for all attributes) at these early-to-mid layers. 

\paragraph{How is attribute knowledge constructed?} We now turn to examine the paths through which we arrive at the final representations $\ell_{attr}$. The overwhelming majority of path lengths (the number of layers in a path, including the embedding layer) are $> 2$, with an average of 5.91 (LLaMA) and 7.97 (Qwen). However, many paths (33.1\% of LLaMA cases and 78.6\% of Qwen's) skip at least one layer, with an average skip size of 0.7 for LLaMA and 2.0 for Qwen (Appendix \ref{app:quantitative_analysis} Figure \ref{fig:skip_ratios}).
For both models, the entities need to be processed by several (but not all) model layers in order for the knowledge to be available to use. The overall number of needed layers is similar for both models, but the Qwen paths are longer and sparser on average.

\paragraph{Existence of alternate paths.}
The alternative path search confirms that models contain multiple, functionally equivalent circuits for factual computation, identifying non-identical alternative paths in 80.1\% (LLaMA) and 82.7\% (Qwen) of cases. Quantitative analysis (Appendix \ref{app:quantitative_analysis}, Figure \ref{fig:experiment2_results}) reveals that alternative paths are universally longer (mean length: 9.47 for LLaMA, 10.47 for Qwen), relying on massive non-sequential jumps (mean skip size: 6.46 layers for LLaMA, 10.4 for Qwen).

\begin{figure}[t]
    \centering
    
    \subfigure[LLaMA 3.1 8B]{
        \includegraphics[width=1.0\linewidth]{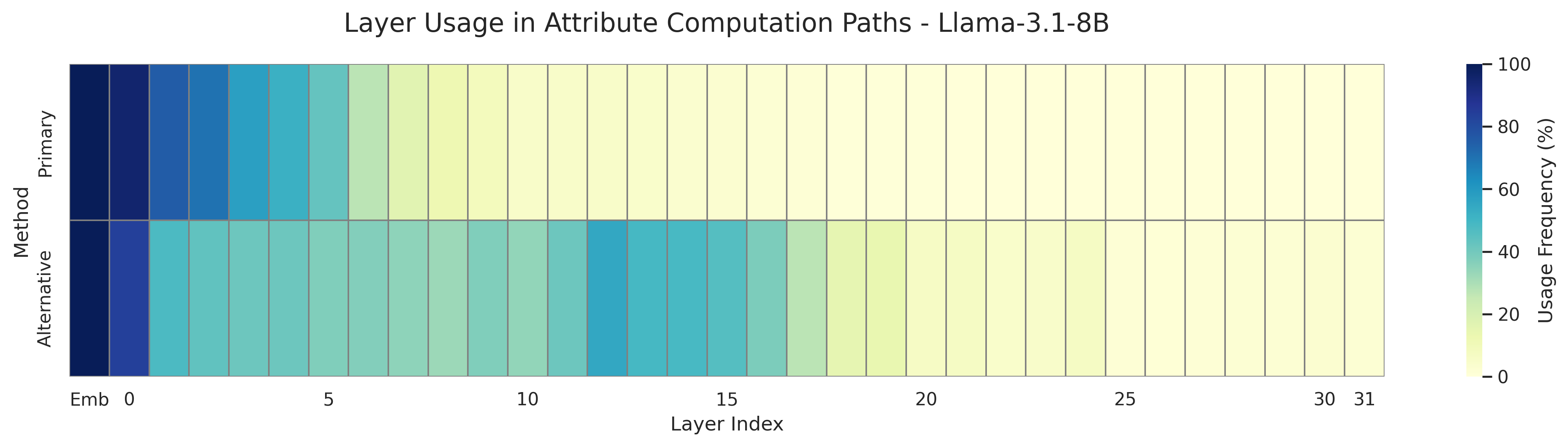}
        \label{fig:layer_use_llama}
    }
    
    \vspace{0.2cm}
    
    \subfigure[Qwen3 8B]{
        \includegraphics[width=1.0\linewidth]{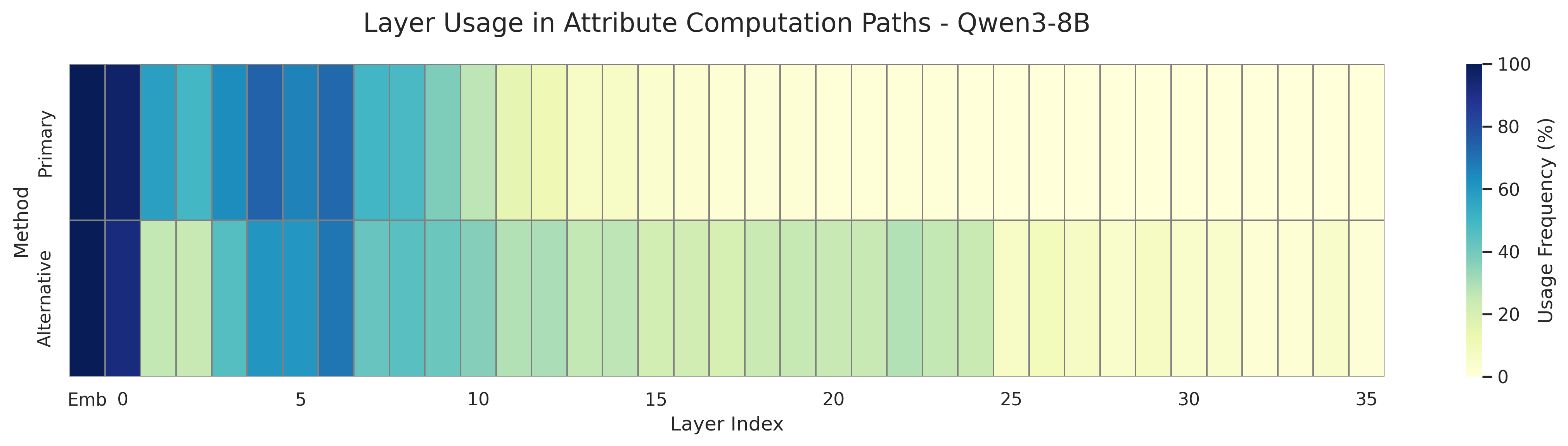}
        \label{fig:layer_use_qwen}
    }
    
    \caption{
        \textbf{Layer Usage Patterns.} Aggregated layer utilization frequency for Primary (top row in each panel) vs. Alternative (bottom row) computation paths.
    }
    \label{fig:layer_usage_heatmap} 
\end{figure}

In addition, $\ell_{attr}$ shifts significantly deeper compared to the primary path—averaging 13.93 for LLaMA and 18.06 for Qwen (Appendix \ref{app:quantitative_analysis}, Figure \ref{fig:experiment1_results}).
Layer usage heatmaps aggregated across paths (Figure \ref{fig:layer_usage_heatmap}) further confirm that these paths recruit a distinct, deeper set of modules, effectively bypassing the primary processing centers. Thus, while redundant paths exist, the primary path consistently offers the most compact route to attribute sufficiency.

\paragraph{Are the constructed representations sufficient?}
To characterize the functional role of $\ell_{attr}$ output ($O^{(e, \ell_{attr})}$), we performed a suite of four targeted patching interventions:
\begin{enumerate}
    \item \textbf{Representation Knockout:} We execute the minimal path up to $\ell_{attr}$, then immediately overwrite the entity representation with a counterfactual one. This tests if the constructed representation in layers above $\ell_{attr}$ is necessary for prediction.
    \item \textbf{Downstream Injection:} We inject the representation from $\ell_{attr}$ into all subsequent layers inputs ($l > \ell_{attr}$) while making all previous layers counterfactual. This tests if the final state alone, without the path's history, is sufficient to drive the upper model layers.
    \item \textbf{Path + Continuation:} We inject the representation from $\ell_{attr}$ into all path layers and to all higher layers ($l > \ell_{attr}$), leaving only off-path intermediate layers as counterfactual.
    \item \textbf{Global Broadcast:} We inject the representation from $\ell_{attr}$ into \textit{all} model layers.
\end{enumerate}

The results of these experiments, detailed in Table \ref{tab:functional_analysis}, yield two insights.
First, the representation at $\ell_{attr}$ is strictly necessary for factual recall; in the Representation Knockout setting, overriding it and following layers with a counterfactual state disrupted the correct prediction in $>94\%$ of cases across both models.
Second, we find that the representation constructed at $\ell_{attr}$ on its own is often not sufficient: while Downstream Injection restores the desired attribute in almost 56\% of primary paths in LLaMA, it fails to do so in the remaining 44\%, for which activations at intermediate path layers are also required.
This indicates that the computation path often acts as a coherent functional unit, where earlier entity representations are used for the final prediction.

\begin{table*}[t]
    \centering
    \small
    \begin{tabular}{lcccc}
        \toprule
        & \multicolumn{2}{c}{\textbf{LLaMA 3.1 8B}} & \multicolumn{2}{c}{\textbf{Qwen3 8B}} \\
        \cmidrule(lr){2-3} \cmidrule(lr){4-5}
        \textbf{Intervention Type} & \textbf{Primary} & \textbf{Alternative} & \textbf{Primary} & \textbf{Alternative} \\
        \midrule
        Representation Knockout & 99.50\% & 94.95\% & 99.40\% & 94.35\% \\
        Downstream Inject & 44.50\% & 76.60\% & 56.05\% & 76.85\% \\
        Path + Cont. & 34.45\% & 48.25\% & 53.20\% & 64.55\% \\
        Global Broadcast & 37.00\% & 61.95\% & 62.65\% & 78.65\% \\
        \bottomrule
    \end{tabular}
    
    \caption{\textbf{Prediction failure rates} across functional analysis experiments. All values indicate the percentage of cases where the model failed to output the correct attribute.}
    \label{tab:functional_analysis}
\end{table*}

\paragraph{The Roles of Path Layers.}
To further investigate the functional role of intermediate representations, we explicitly consider the two ways in which path layers can be utilized. A path layer can either act as an intermediate transformation—producing representations that will be fed to subsequent layers at the same token position—or it can enable inter-token transfer, moving information about the entity from the layer's input to subsequent sequence positions via the attention mechanism. To determine which function a layer serves, we test each identified path (both primary and alternative) using an extended \textit{isolate} operation. Iteratively, we feed each path layer counterfactual \emph{inputs} while restoring its entity-position \emph{outputs} to their clean-path states. If the final prediction remains correct, the layer is solely performing same-token representation transformation, rather than propagating necessary information to later sequence positions. For these layers, we leave this transformation-only restriction in place while testing higher layers. We find that reliance on this propagation correlates strongly with path length: primary paths require at least one layer to act as an inter-token propagator in 50.1\% of cases (average path length 7.4, vs. 4.4 when not required), while alternative paths require it in 80.6\% of cases (average length 10.8, vs. 3.9).
Similar patterns are observed in Qwen3 8B. For primary paths, 63.8\% require information propagation (average path length 8.64, vs. 6.78 when not required). In alternative paths, this dependency increases to 82.4\% (average path length 11.11, vs. 7.49).
This indicates that longer, non-standard computation paths depend more heavily on information propagation across the sequence.
Notably, approximately 15\% of all paths across both Llama and Qwen require clean information propagation as early as Layer 0. This is unexpected under our hypothesis that the specific attribute would not yet be computed at this initial stage. Detailed layer-wise usage for information propagation is visualized in the heatmaps in Appendix \ref{app:info_prop}.

\paragraph{Constructing additional paths via recombination.}
We test the viability of ``hybrid'' paths composed of early steps from a primary path and later steps from an alternative path, by concatenating a prefix from a primary path with a valid suffix from the corresponding alternative path, enforcing that the resulting path is distinct from the originals and non-trivial (the chosen prefix from the primary does not contain the corresponding initial segment of the alternative). Each hybrid is evaluated using our counterfactual patching method (\autoref{subsec:path-ident}); success is defined as restoring the target prediction and achieving attribute sufficiency at the final layer.

The results reveal significant modular flexibility. In LLaMA 3.1 8B, 35.4\% of prompts admit at least one successful hybrid path (avg. 2.75 successful combinations per prompt), while Qwen3 8B shows a 32.8\% success rate with higher redundancy (avg. 4.13 combinations). This proves the existence of additional computation paths beyond those identified in previous experiments, demonstrating that different layer sequences can perform functionally equivalent transformations. It also indicates that layer roles in the primary and alternate paths do not necessarily map one-to-one: in some cases a single layer in the primary path is accounted for by several layers in the alternate path, or vice versa.

\subsection{Additional Experiments and Analyses}
\label{subsec:additional_exp}
\paragraph{Effect of order.} While in most prompts the relation appears before the entity (\textit{``The mother tongue of X is''}), in few of them the order is reversed (\textit{``X's mother tongue is''}). Does this affect the retrieval process? We could not identify a strong trend, beyond the paths concluding on later layers in LLaMa (but not in Qwen). However, this could be due to small sample size, and could be the topic of a future study. Further details are available in Appendix \ref{sec:appendix_position}.

\paragraph{Paths in Individual Relations.} 
We analyzed relation types with sufficient coverage ($N \ge 50$). While not a universal trend, we observe anecdotal examples suggesting a potential link between semantic specificity and computational cost (Appendix \ref{app:rel_analysis}). In certain instances, broad categorical relations (e.g., \textit{Continent}) require fewer layers than more specific factual retrievals (e.g., \textit{Headquarters}), hinting that the model might occasionally recruit longer circuits to resolve highly specific attributes. Additionally, per-relation layer-usage heatmaps in Appendix \ref{app:rel_analysis} illustrate that distinct relations recruit specific sets of layers beyond simple path length differences.
However, this length hierarchy is highly inconsistent: we observe significant rank discordance between models and between primary and alternative paths. The fact that relative difficulty shifts in the alternative setting implies that ``backup'' mechanisms utilize distinct processing logic rather than simply mirroring primary circuits.

\paragraph{Relation to Entity Resolution.}
Entity resolution (ER) refers to the stage in which a model forms an internal entity representation sufficient to explicitly reconstruct the entity's name within a suitable target context \citep{ghandeharioun2024patchscopes}. To examine the relationship between ER and attribute computation, we apply the Patchscopes method \citep{ghandeharioun2024patchscopes} to \emph{multi-token} entities along our identified paths ($N_{LLaMA}=1,972$, $N_{Qwen}=1,945$).
We use the original constant prompt from this method, of the form ``$e_1: d_1$,\ldots, $e_k: d_k$, $x$'' where each description $d_i$ begins with the entity name $e_i$. We extract the entity's final token representation from every layer along its computation path (for both primary and alternative paths) and inject it into the $x$ position across all target layers. We then identify the first source--target layer pair, if any, that decodes the full multi-token entity name.

The results (Appendix \ref{app:entity_resolution}) reveal a strong dissociation between ER and attribute sufficiency. Explicit ER steps are detected in a minority of primary paths (12.9\% for LLaMA, 9.1\% for Qwen) and even fewer alternative paths (9.1\% and 6.6\%). This implies that detectable ER is not a universal prerequisite for factual recall; while primary paths retain slightly stronger identity traces, attribute retrieval often occurs implicitly without a distinguishable ``identity state'' accessible to Patchscopes.

\paragraph{Impact of Counterfactual Noise on Attribute Computation.}
To test whether counterfactual noise in the \emph{isolate} operation artificially alters identified paths—due to incorrect information propagating through non-path layers—we additionally experiment with a modified algorithm: instead of injecting counterfactual activations into excluded layers, we supply them with the most recent preceding path layer's input (effectively ``locking'' the representation). 
Under this setting, average path lengths remain highly consistent with the original operation across both LLaMA and Qwen (noise-free vs. original — LLaMA primary: 5.88 vs. 5.91, alternative: 8.99 vs. 9.47; Qwen primary: 7.92 vs. 7.97, alternative: 10.69 vs. 10.47). Primary path termination depth (mean $\ell_{attr}$) is also virtually identical for both models (LLaMA: 4.65 vs. 4.61; Qwen: 8.02 vs. 7.97)—an expected outcome since $\ell_{attr}$ serves as the explicit search bound. However, while alternative path lengths remain stable, their computation terminates deeper in this locked setting (LLaMA: mean layer 17.73 vs. 13.93; Qwen: 20.87 vs. 18.06). Layer usage heatmaps (Appendix \ref{app:noise_free_heatmaps}) confirm this shift is driven by increased utilization of the deepest layers in both models; LLaMA's usage of layers 27–30 jumps from 1–3\% to 9–15\%, and Qwen exhibits a similar late-stage spike, peaking at 22\% in its penultimate layer (layer 34).
We hypothesize that counterfactual noise ``forces'' the model to compute the attribute earlier to override conflicting signals; without this noise, our greedy search naturally defers computation. Overall, while the \emph{amount} of required computation (path length) is an intrinsic property of the task, the \emph{depth} of this extraction is highly sensitive to counterfactual noise.

\section{Discussion and Conclusion}

We introduced the concept of an Attribute-Computation Path and a novel patching protocol to identify such paths in LLMs. Our analysis confirmed that the computation of an attribute is a multi-step, non-contiguous process that frequently skips layers and requires multiple necessary steps. Furthermore, we show that models possess multiple, functionally equivalent computation paths for the same fact, emphasizing a high degree of computational redundancy. 

Our findings suggests that the dominant folk view of factual storage and retrieval, in which facts are stored in specific MLP layers, is simplistic and misleading: the reality appears to be significantly more complex, with knowledge being stored in a distributed way across multiple layers, that needs to be processed together for an effective recall of an individual fact.

\section*{Limitations}

While we showed redundancy in minimal attribute computation paths exists, we did not fully quantify it. Our estimation of model redundancy is constrained by our search strategy in two primary ways. First, our reliance on an iterative greedy search algorithm—rather than an exhaustive combinatorial search—means that we naturally may miss many valid computational paths that might arise from different layer combinations. Second, we explicitly targeted two distinct path types by defining specific upper bounds for this search ($\ell_{attr}$ and the final model layer). Consequently, our findings likely represent only \emph{a lower bound} on the true degree of redundancy; it is plausible that a granular sweep of search bounds across all intermediate layers would reveal a much larger spectrum of functionally equivalent paths that our current analysis did not explore.

Second, while our interventions demonstrate that these redundant paths can be mechanistically accessed in an intervention, it remains unclear whether the model utilizes these deeper layers for entity enrichment during a standard, unperturbed forward pass, or if they represent degenerate remnants of the training process.

\bibliography{references}

\clearpage
\appendix

\section{Visualization of $\ell_{attr}$ Identification}
\label{app:visualizations}

\begin{figure*}[h]
    \centering
    \includegraphics[width=\textwidth]{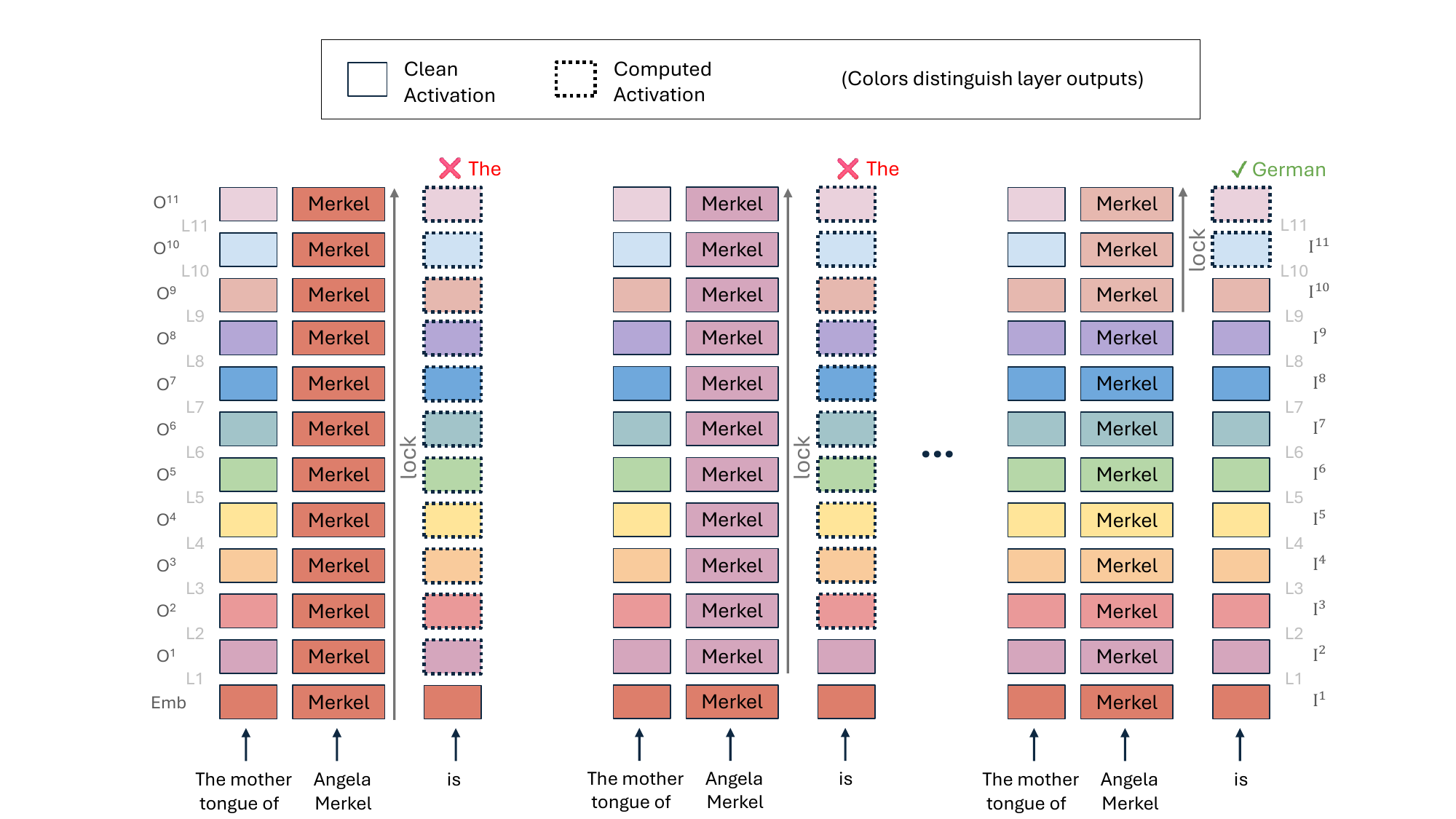}
    \caption{
        \textbf{$\ell_{attr}$ Identification.} 
        This process identifies $\ell_{attr}$, the first layer where the entity representation is robust enough to elicit the target attribute without further processing.
        The labeled squares represent the entity representation at different layers.
        \textbf{Left \& Middle (Testing an Insufficient Layer):} We lock the ``Angela Merkel'' representation at an early layer. The model fails to retrieve the correct attribute, predicting a generic or nonsensical token (e.g., \textit{``The''}), indicating the representation is not yet sufficient.
        \textbf{Right (Testing a Sufficient Layer):} We lock the representation at a deeper layer. The model successfully predicts \textit{``German''}, identifying this layer as $\ell_{attr}$.
    }
    \label{fig:l_attr_visualization}
\end{figure*}

Figure~\ref{fig:l_attr_visualization} provides a detailed visualization of the $\ell_{attr}$ identification process, using the prompt \textit{``The mother tongue of Angela Merkel is''} (Target: \textit{``German''}). The figure illustrates this step-by-step procedure in which we lock the entity representation at each layer—starting from the embeddings and moving upwards—until the operation successfully reproduces the correct target token.

\section{Extended Quantitative Analysis}
\label{app:quantitative_analysis}

\begin{figure*}[t!]
    \centering
    
    \subfigure[Distribution of $\ell_{attr}$ (primary method).]{
        \includegraphics[width=0.48\textwidth]{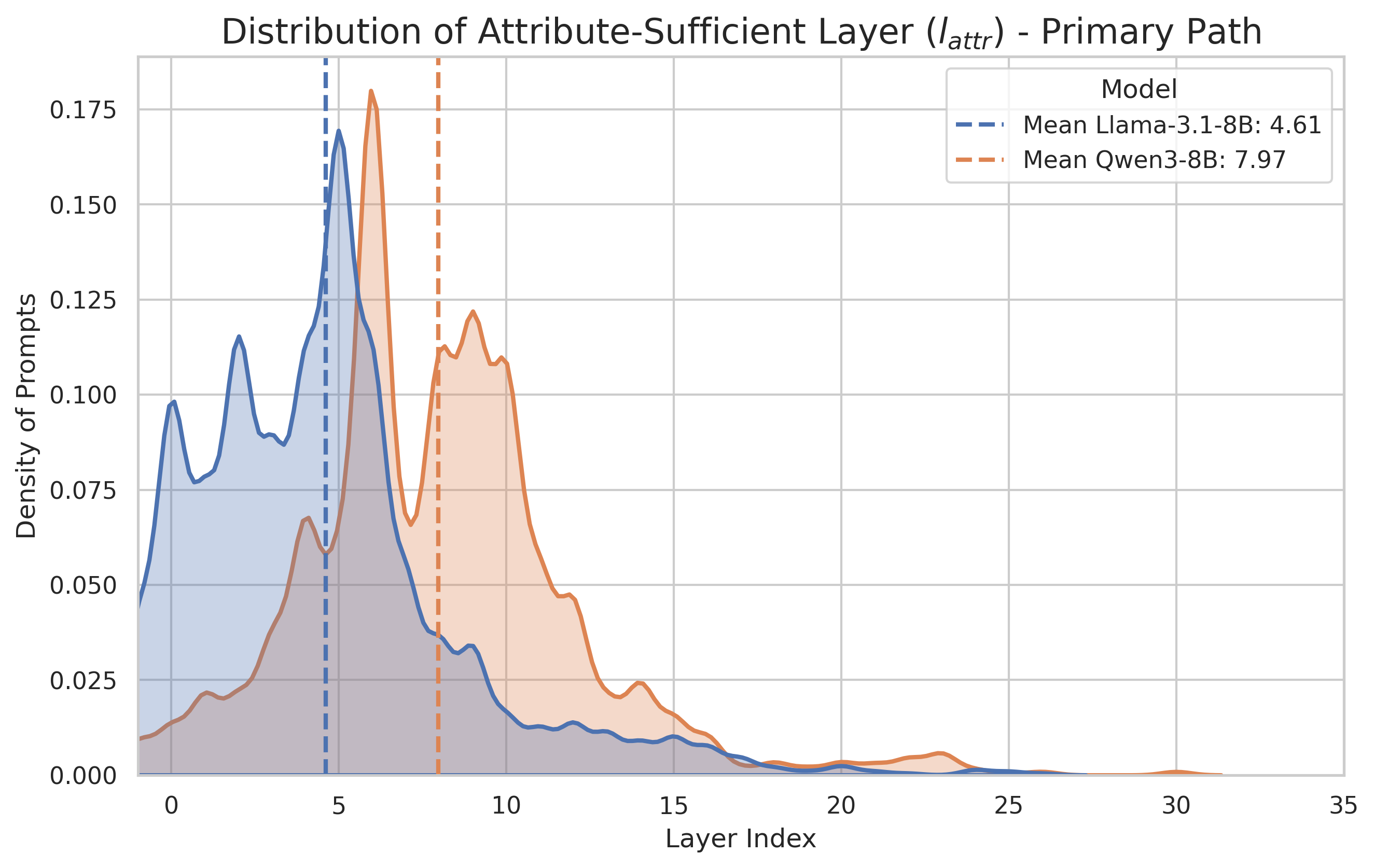}
        \label{fig:l_attr_dist}
    }
    \hfill
    \subfigure[Distribution of $\ell_{attr}$ (Alternative Path)]{
        \includegraphics[width=0.48\textwidth]{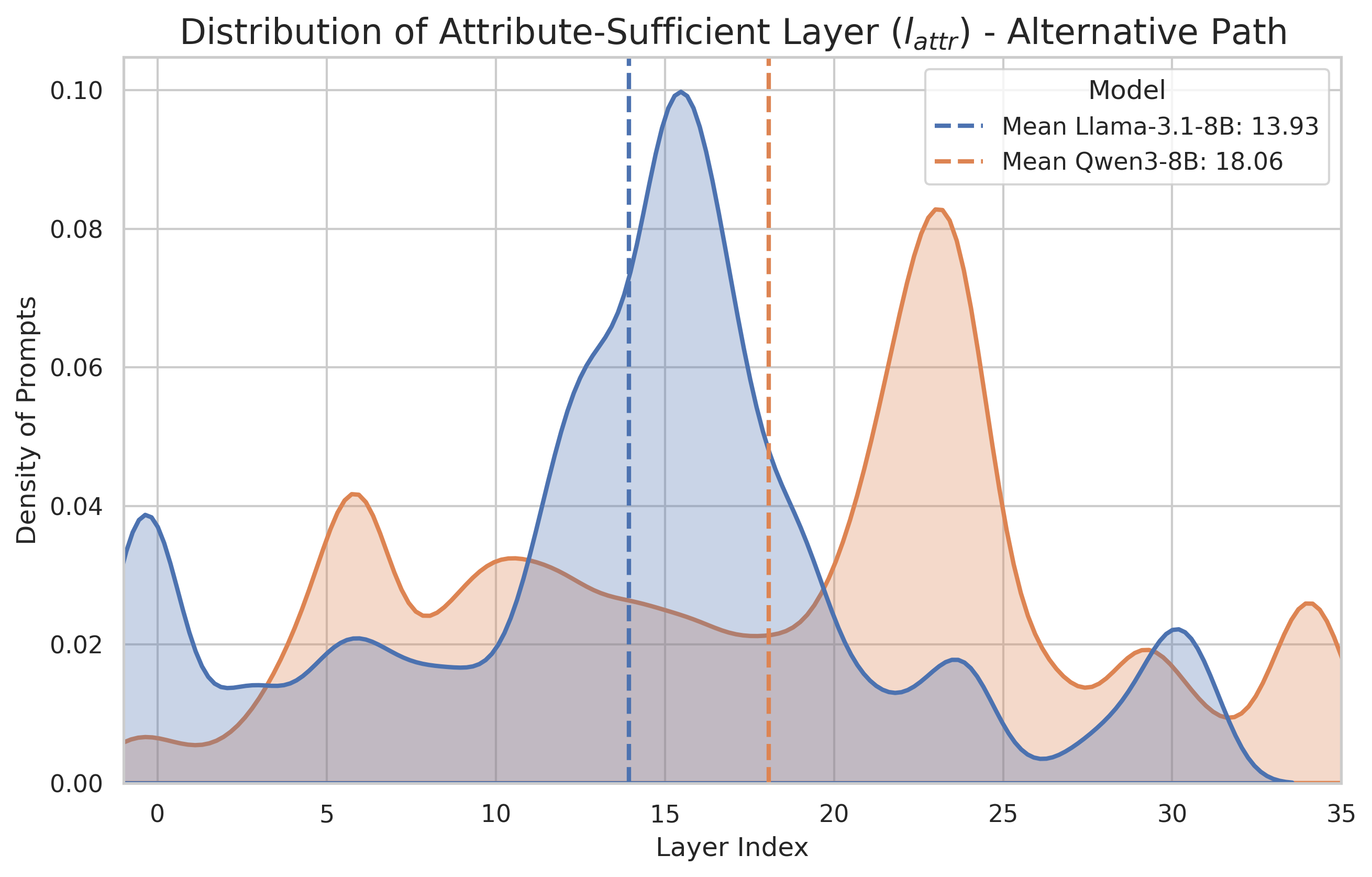}
        \label{fig:l_attr_dist_alt}
    }
    
    \caption{
        Analysis of minimal computation paths for the primary method.
        (a) The distribution of $\ell_{attr}$, showing both models complete attribute computation in the early-to-mid layers.
        (b) The distribution of the final $\ell_{attr}$ for the alternative path, which is significantly deeper and wider than the primary path's distribution.
    }
    \label{fig:experiment1_results}
\end{figure*}

\begin{figure*}[t!]
    \centering
    
    \subfigure[$\ell_{attr}$: Primary vs. Alternative Path]{
        \includegraphics[width=0.48\textwidth]{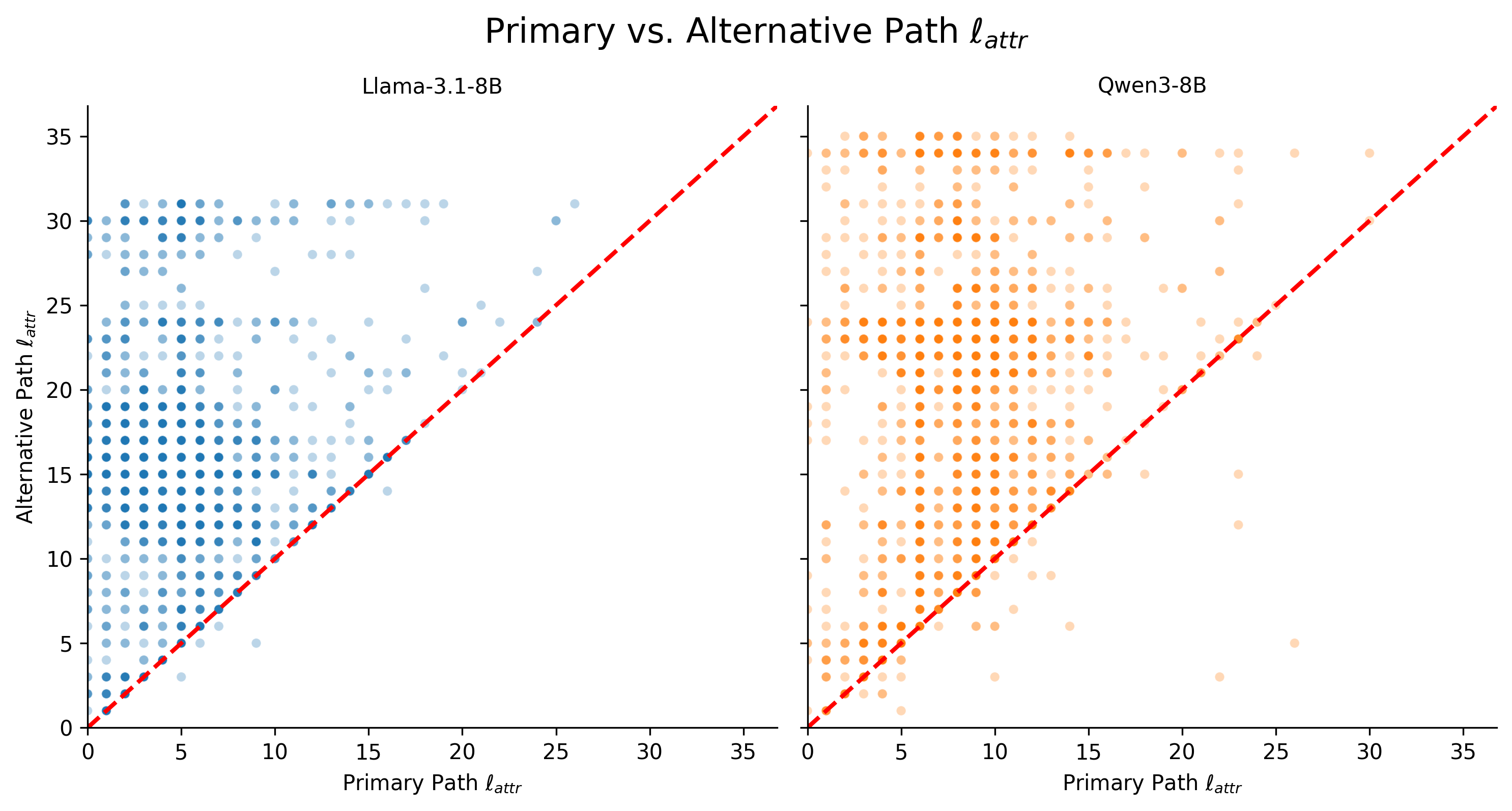}
        \label{fig:l_attr_pvc}
    }
    \hfill
    \subfigure[Path Length: Primary vs. Alternative Path]{
        \includegraphics[width=0.48\textwidth]{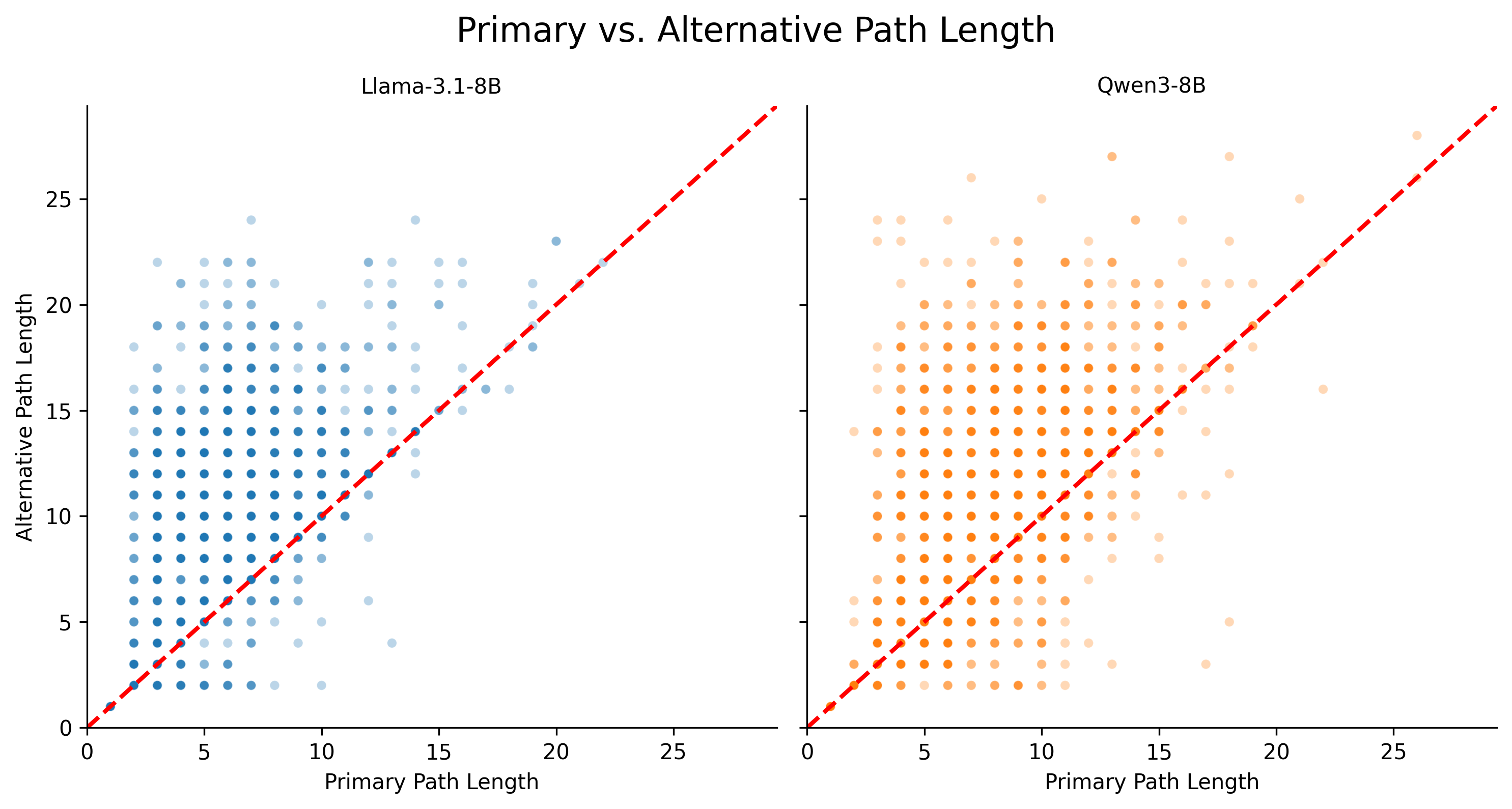}
        \label{fig:path_len_pvc}
    }

    \vspace{0.5cm}
    \subfigure[Distribution of Path Compression Ratios (LLaMA).]{
        \includegraphics[width=0.48\textwidth]{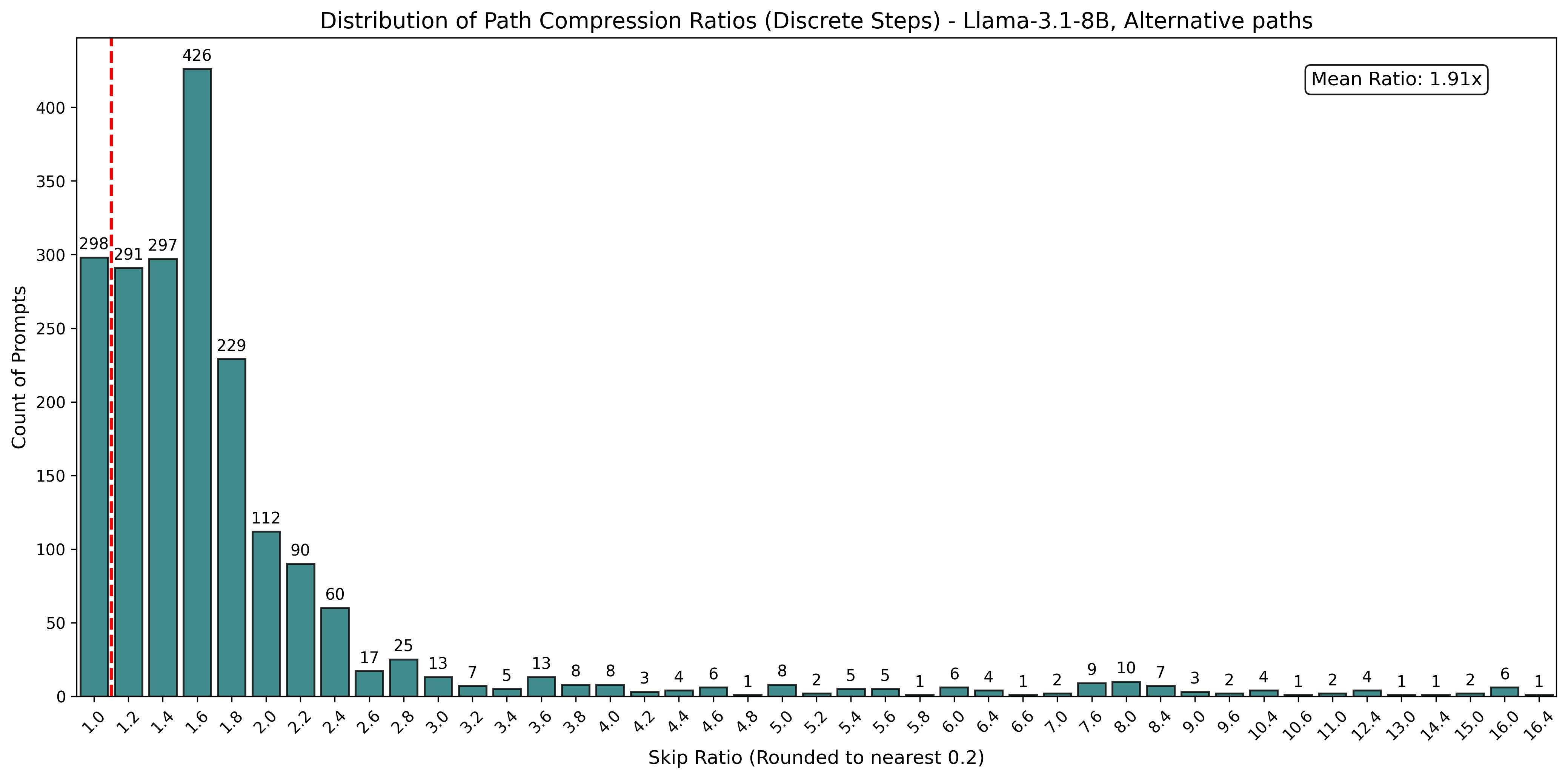}
        \label{fig:skip_ratio_llama}
    }
    \hfill
    \subfigure[Distribution of Path Compression Ratios (Qwen)]{
        \includegraphics[width=0.48\textwidth]{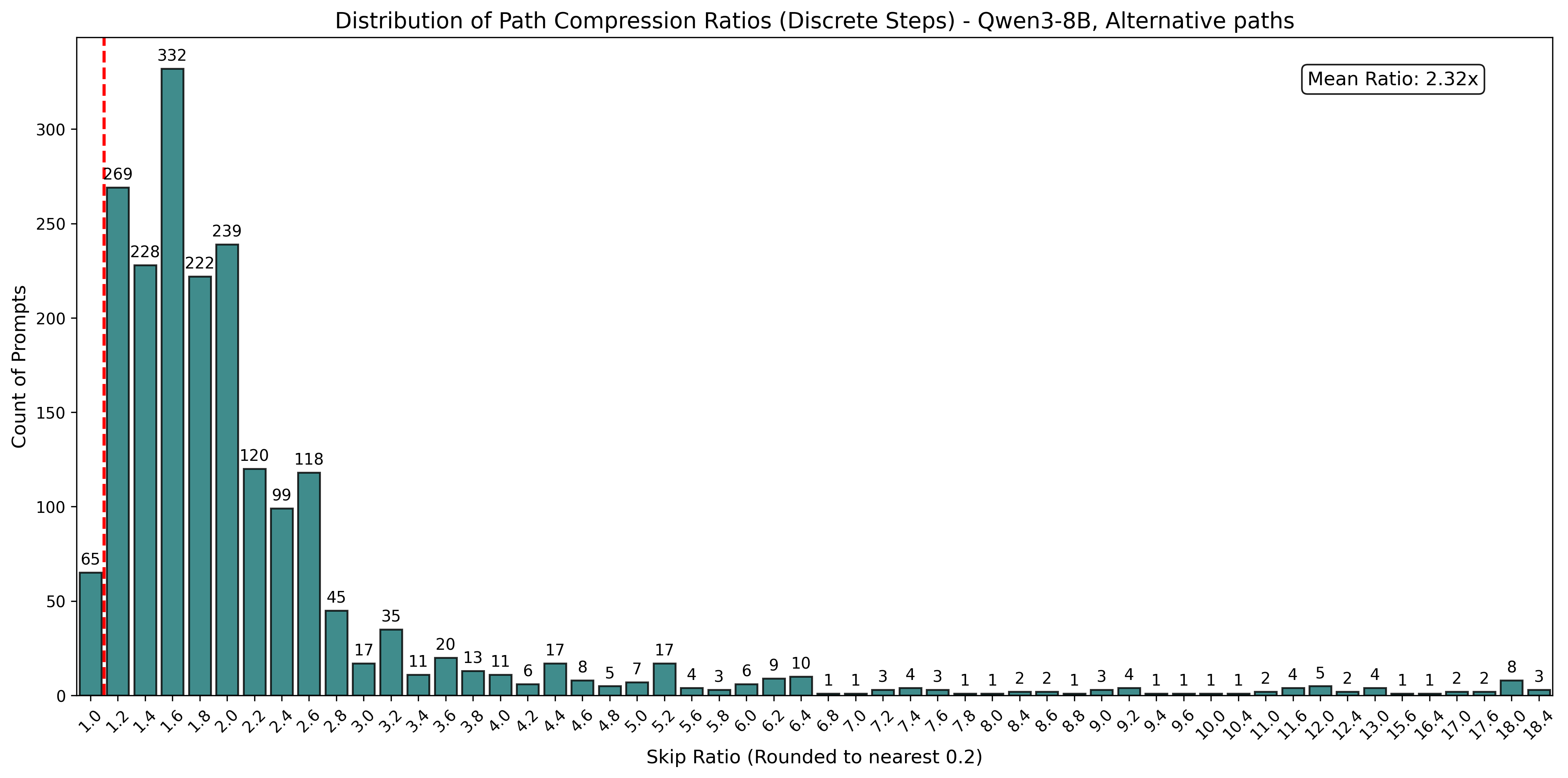}
        \label{fig:skip_ratio_qwen}
    }
    
    \caption{
        Analysis of the alternative computation path.
        (a) Comparison of $\ell_{attr}$ for the primary path (x-axis) vs. the alternative path (y-axis). Most points lie above the $y=x$ line, showing the alternative path is deeper.
        (b) Comparison of path length. Points are again mostly above the $y=x$ line, indicating the alternative path is also longer (more computational steps).
        (c) Distribution of Path Compression Ratios for LLaMA 3.1 8B. The ratio is computed as the path depth divided by the number of steps. The first bar (ratio=$1.0$) represents purely sequential paths. Ratios $>1.0$ are rounded to the nearest $0.2$ for visualization.
        (d) The corresponding distribution for Qwen3 8B.
    }
    \label{fig:experiment2_results}
\end{figure*}

In this section, we provide the detailed statistical breakdown of the identified paths. Figure \ref{fig:experiment1_results} illustrates the distribution of the $\ell_{attr}$ for both the primary and alternative strategies, as well as the layers skipping ratio in the paths. Figure \ref{fig:experiment2_results} offers a direct comparison between the primary and alternative paths, quantifying the increase in depth and path length when the model is forced to utilize later layers for attribute computation.

\begin{figure*}[h]
    \centering
    
    \subfigure[Distribution of Path Compression Ratios (LLaMA).]{
        \includegraphics[width=0.48\textwidth]{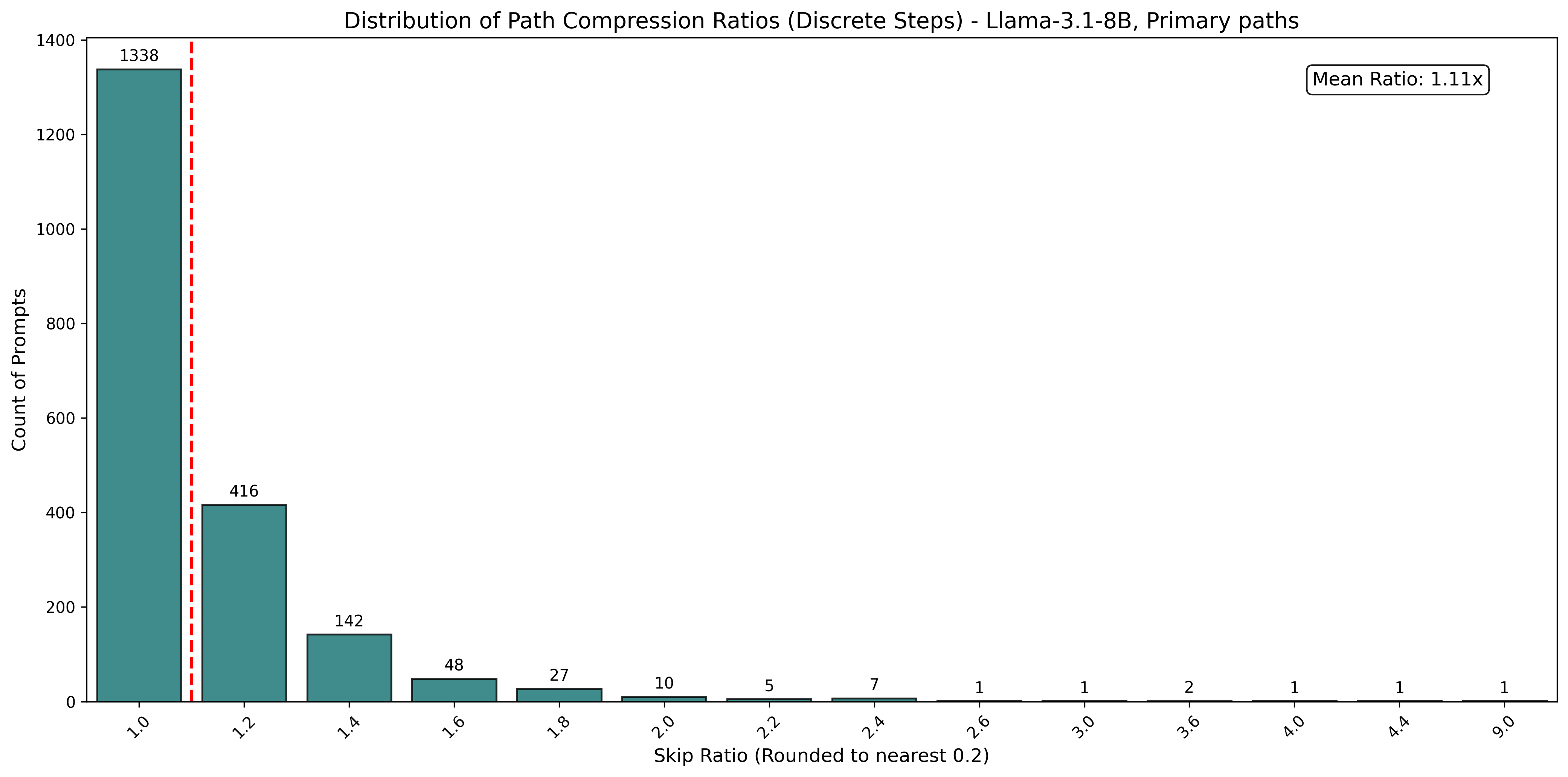}
        \label{fig:skip_ratio_llama}
    }
    \hfill
    \subfigure[Distribution of Path Compression Ratios (Qwen)]{
        \includegraphics[width=0.48\textwidth]{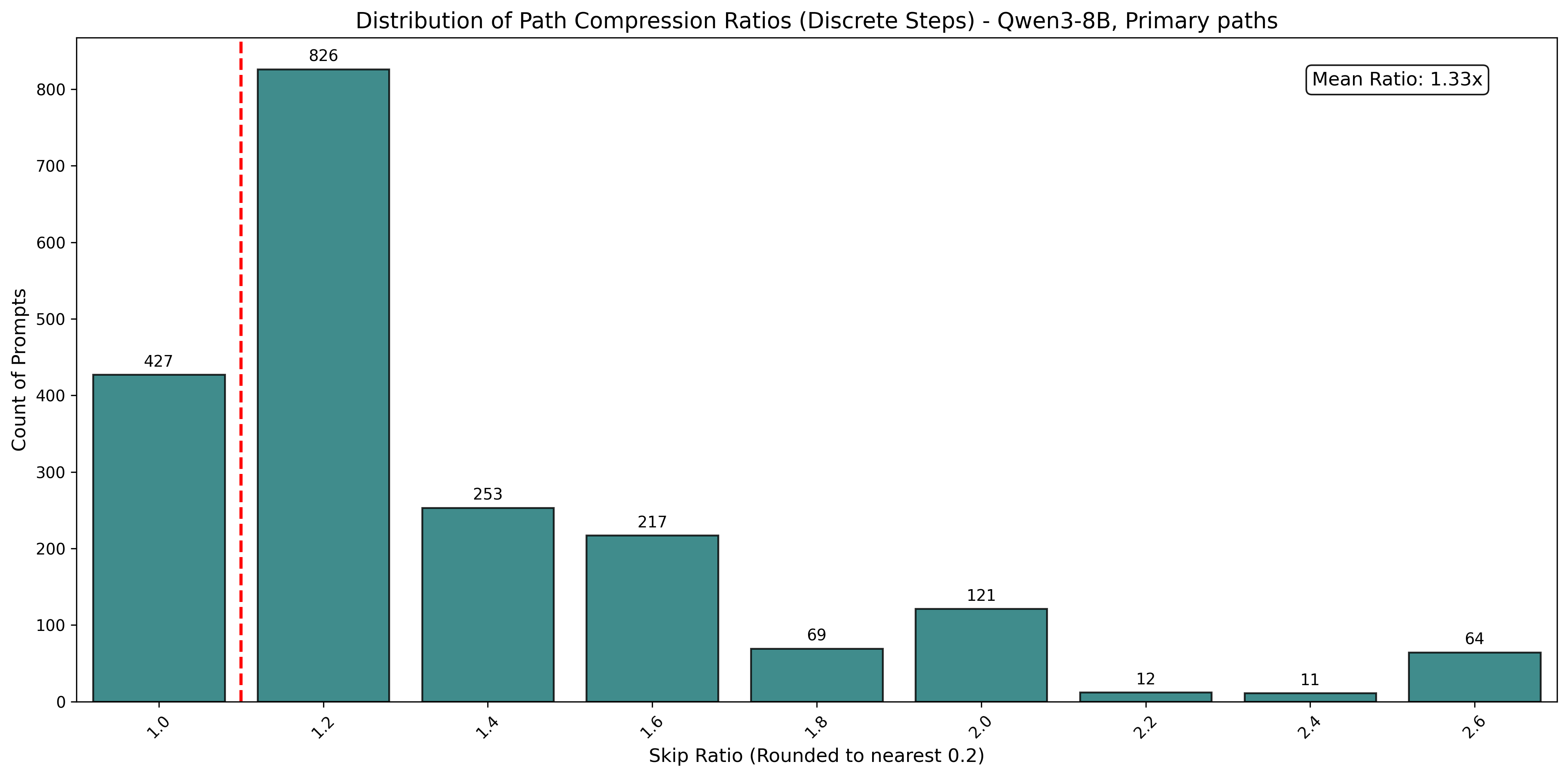}
        \label{fig:skip_ratio_qwen}
    }
    
    \vspace{0.5cm}
    
    \subfigure[Path length vs. path span (primary method).]{
        \includegraphics[width=0.7\textwidth]{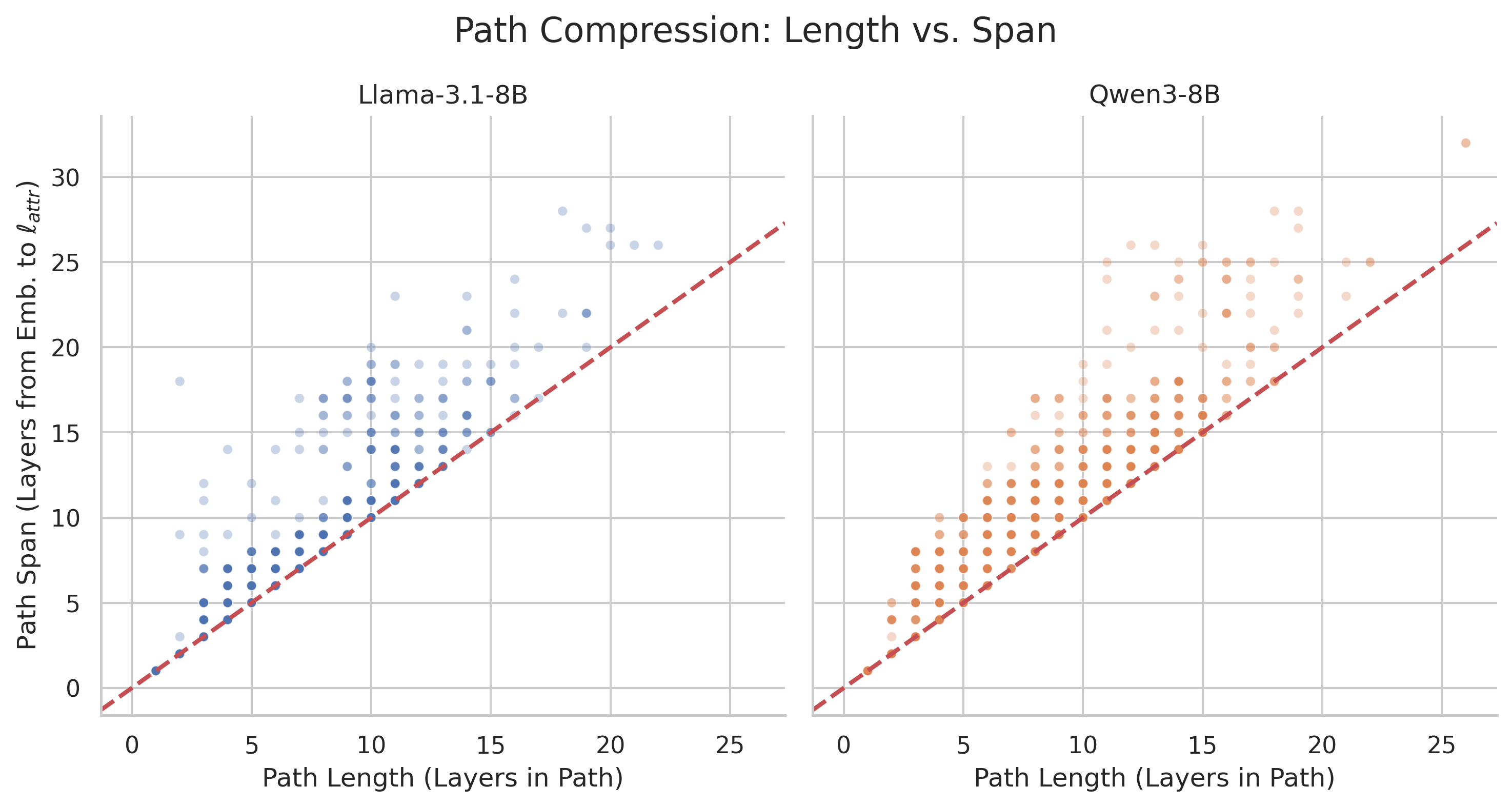}
        \label{fig:path_compression}
    }

    \caption{
        Analysis of minimal computation paths for the primary method.
        (a) Distribution of Path Compression Ratios for LLaMA 3.1 8B. The ratio is computed as the path depth divided by the number of steps. The first bar (ratio=$1.0$) represents purely sequential paths. The red dashed line acts as a boundary, separating these sequential paths from the compressed paths (ratio $> 1.0$) to the right. Ratios $>1.0$ are rounded to the nearest $0.2$ for visualization.
        (b) The corresponding distribution for Qwen3 8B.
        (c) A comparison of actual path length vs. path span for the primary paths. Dots below the $y=x$ line represent ``compressed'' paths that skip layers.
    }
    \label{fig:skip_ratios}
\end{figure*}

\section{Necessary Information Propagation}
\label{app:info_prop}
In this section, we provide a detailed breakdown of layer usage across the computation paths identified in our experiments. We compare the Original Path distribution (where all layers are active) with the Necessary Attention distribution. The latter represents only the layers where intra-layer information propagation was strictly required to maintain the correct prediction.

\begin{figure}[ht]
    \centering

    \subfigure[LLaMA 3.1 8B - Primary Paths]{
        \includegraphics[width=1.0\columnwidth]{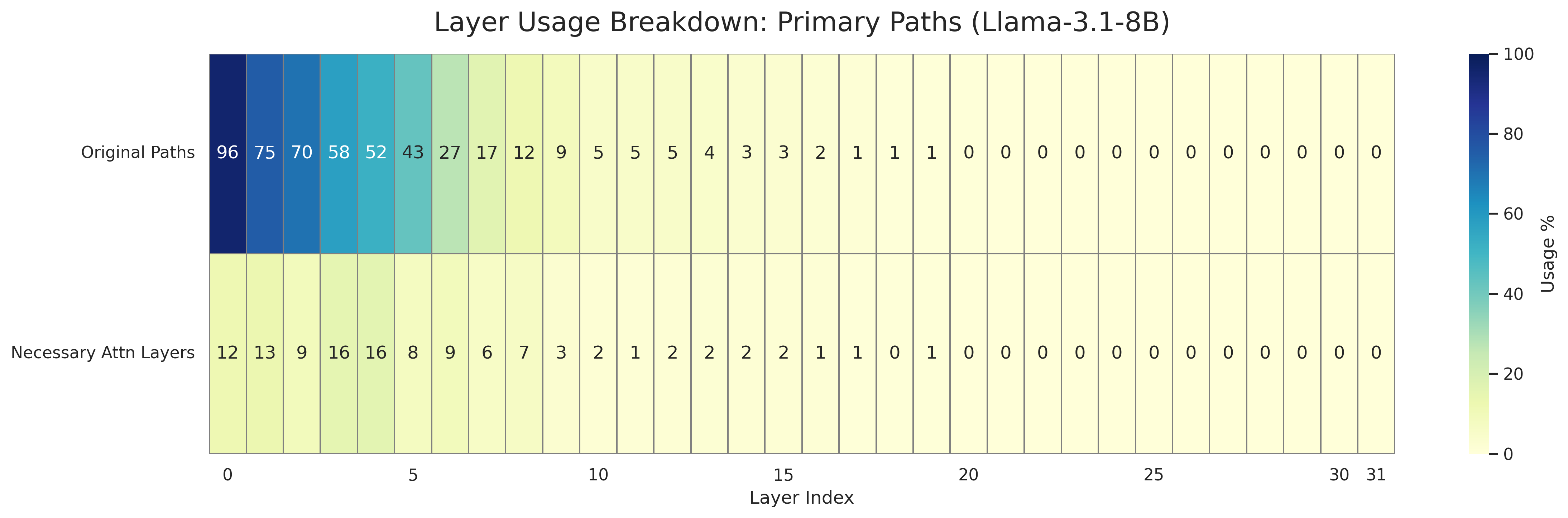}
    }
    
    \vspace{0.3cm}

    \subfigure[LLaMA 3.1 8B - Alternative Paths]{
        \includegraphics[width=1.0\columnwidth]{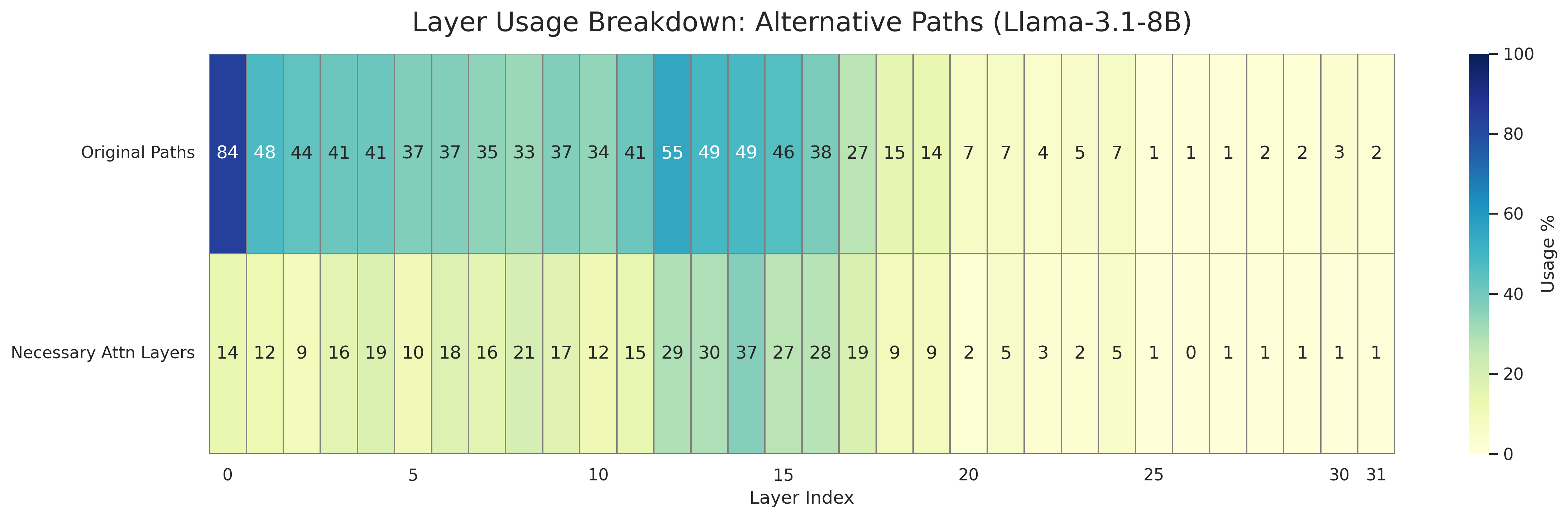}
    }

    \vspace{0.3cm}

    \subfigure[Qwen3 8B - Primary Paths]{
        \includegraphics[width=1.0\columnwidth]{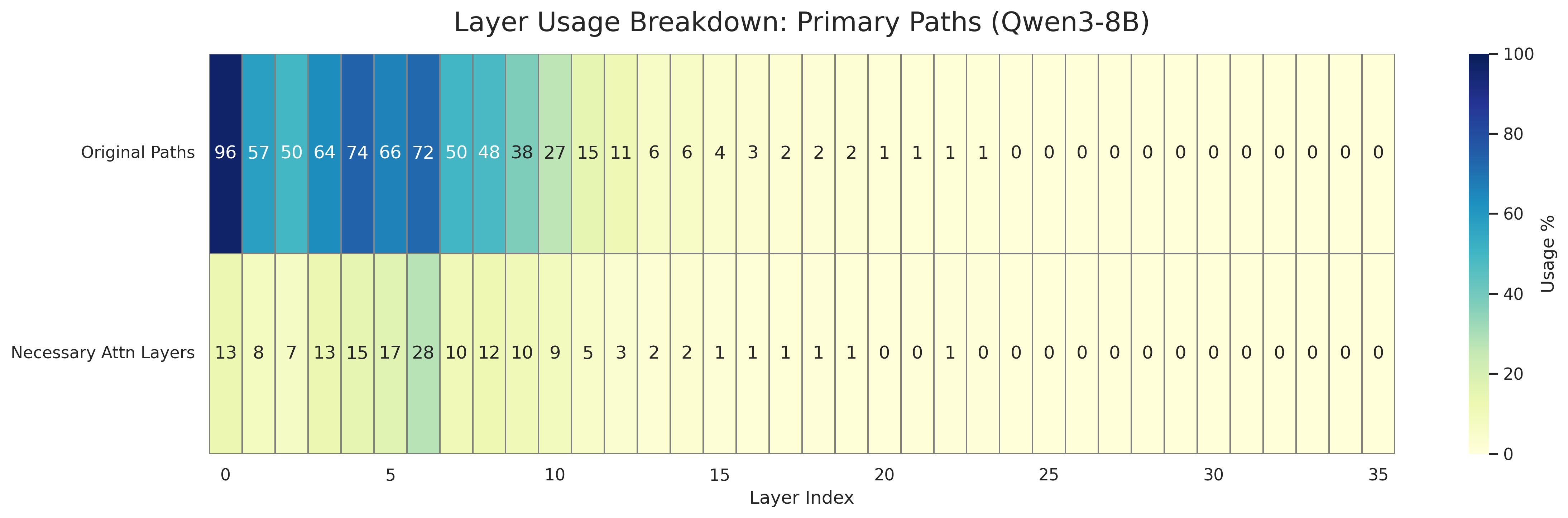}
    }
    
    \vspace{0.3cm}
    
    \subfigure[Qwen3 8B - Alternative Paths]{
        \includegraphics[width=1.0\columnwidth]{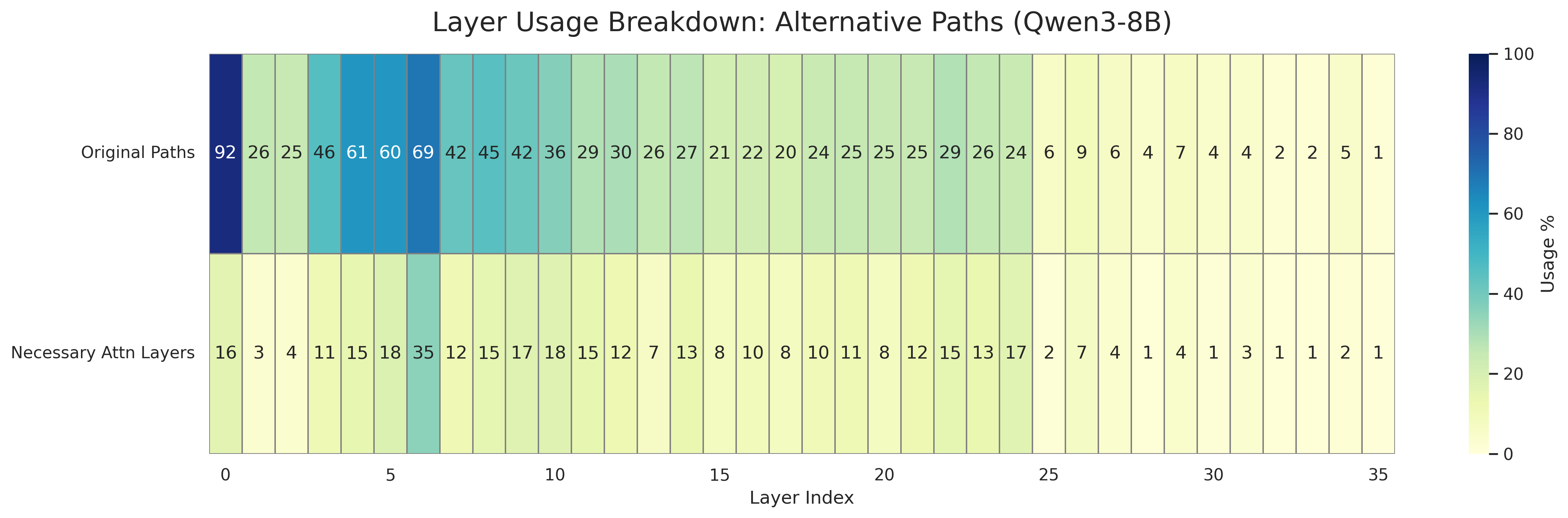}
    }
    \caption{\textbf{Information Propagation Heatmaps.} Each plot compares the percentage of paths utilizing a specific layer in the original identified path (Top Row) versus the subset of layers where clean information propagation was found to be necessary (Bottom Row). Note the non-zero usage at Layer 0 across all configurations.}
    \label{fig:heatmaps_attn}
\end{figure}

The heatmaps in Figure \ref{fig:heatmaps_attn} illustrate the frequency (as a percentage of total paths) with which each layer is utilized. We highlight two key observations:

Correlation with Depth: Consistent with our main findings, longer alternative paths show a higher density of necessary attention layers in the middle and late stages of the model.

The Layer 0 Surprise: Across both Llama and Qwen, approximately 15\% of paths require clean information propagation at Layer 0. This contradicts the initial hypothesis that attribute-specific information is only computed in later blocks.

\section{Analysis by Prompt Structure (Entity Position)}
\label{sec:appendix_position}

Here, we provide detailed results of the experiments. First, we classified each prompt as one of the following types:
\begin{itemize}
    \item \textbf{entity\_first}: The entity appears before the relation (e.g., ``Albert Einstein was born in...'').
    \item \textbf{relation\_first}: The relation appears before the entity (e.g., ``The birthplace of Albert Einstein is...'').
\end{itemize}

To perform this classification, we used GPT-4o. We provided the model with the formatted prompt (e.g., ``The official religion of Edwin of Northumbria is'') and requested a classification using the few-shot prompt structure in Figure \ref{fig:prompt_template}.

\begin{figure*}[h] 
    \centering
    \begin{verbatim}
    Example 1:
    Sentence: The official religion of Edwin of Northumbria is
    Answer: relation
    
    Example 2:
    Sentence: In Nykarleby, the language spoken is
    Answer: entity
    
    Now, analyze the following sentence.
    What appears first: the relation or the entity?
    
    Sentence: {prompt}
    Answer with only 'relation' or 'entity'.
    \end{verbatim}
    \caption{The few-shot prompt used to classify the ordering of entity and relation in the query.}
    \label{fig:prompt_template}
\end{figure*}

Figure \ref{fig:prompt_struct} shows the comparison of $\ell_{attr}$ and path length between these two prompt types for the primary and the alternative paths.
We restricted our comparison to relations that had at least 50 samples for both structure types together across both models.
The number above each bar indicates the sample size used to calculate the average.

\begin{figure*}[p]
    \centering

    \subfigure[$\ell_{attr}$ vs. Entity Position - LLaMA 3.1 8B.]{
        \includegraphics[width=0.8\textwidth]{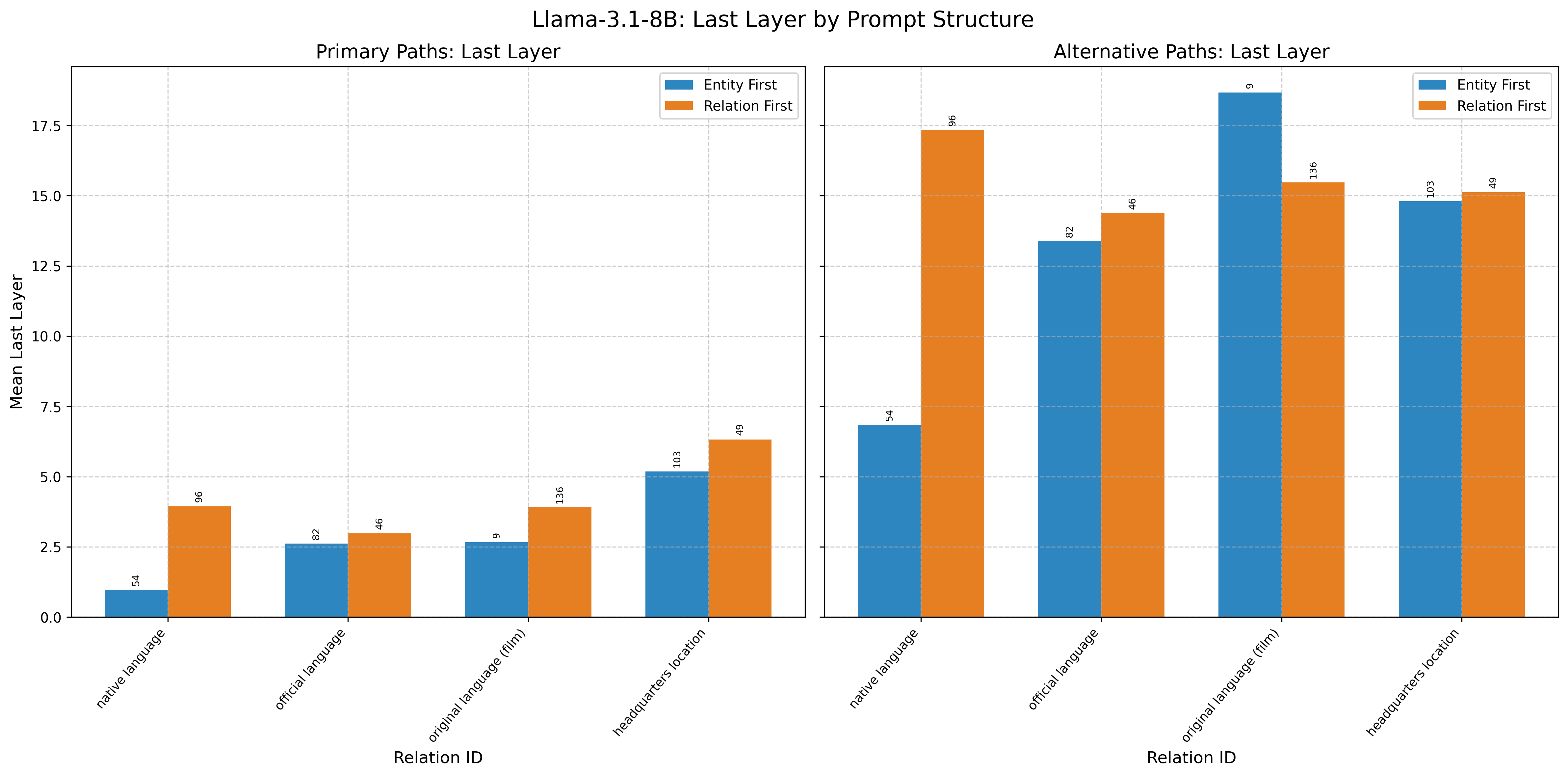}
    }
    
    \vspace{0.5cm}

    \subfigure[Path Length vs. Entity Position - LLaMA 3.1 8B.]{
        \includegraphics[width=0.8\textwidth]{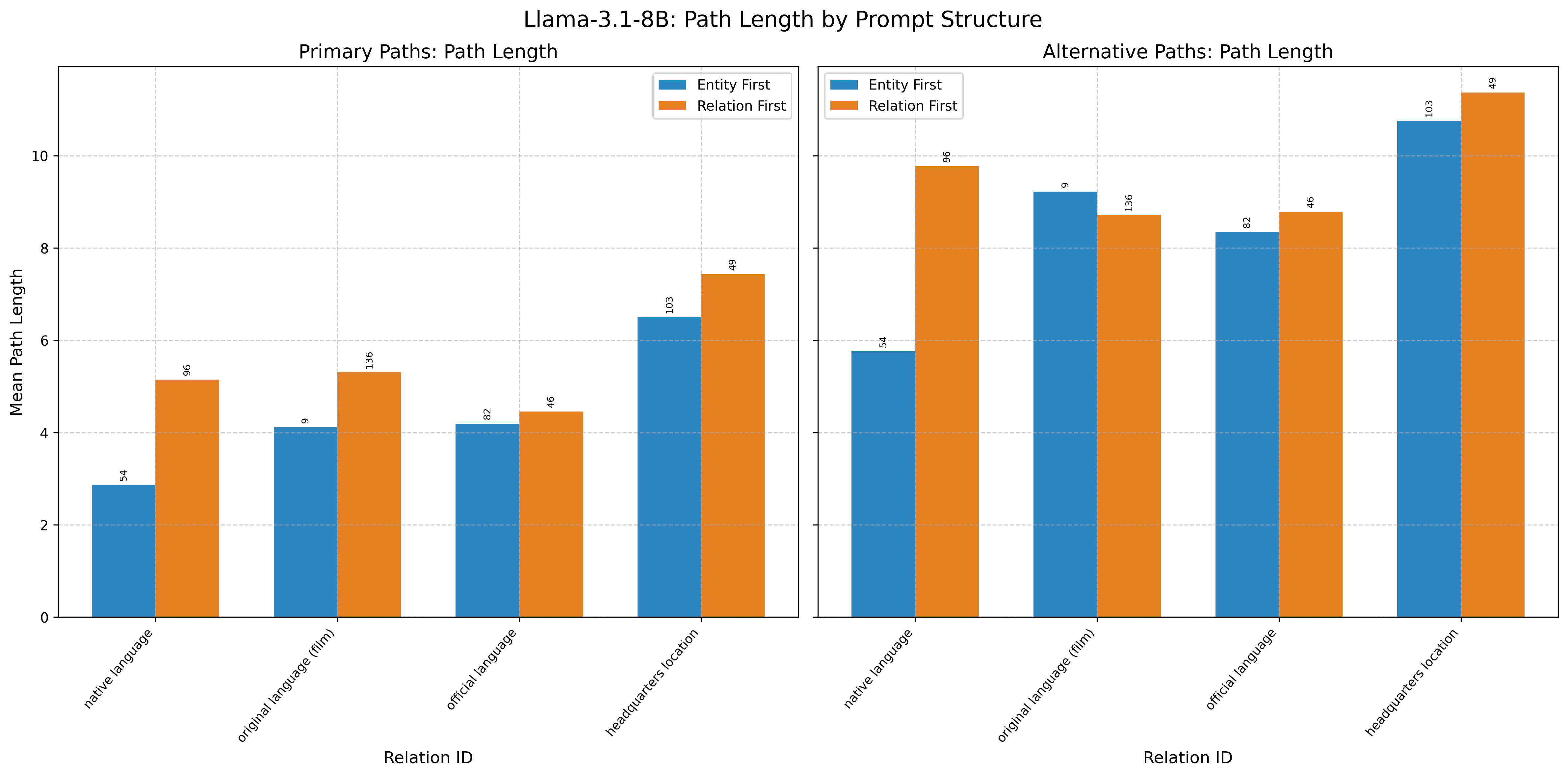}
    }

    \vspace{0.5cm}

    \subfigure[$\ell_{attr}$ vs. Entity Position - Qwen3 8B.]{
        \includegraphics[width=0.8\textwidth]{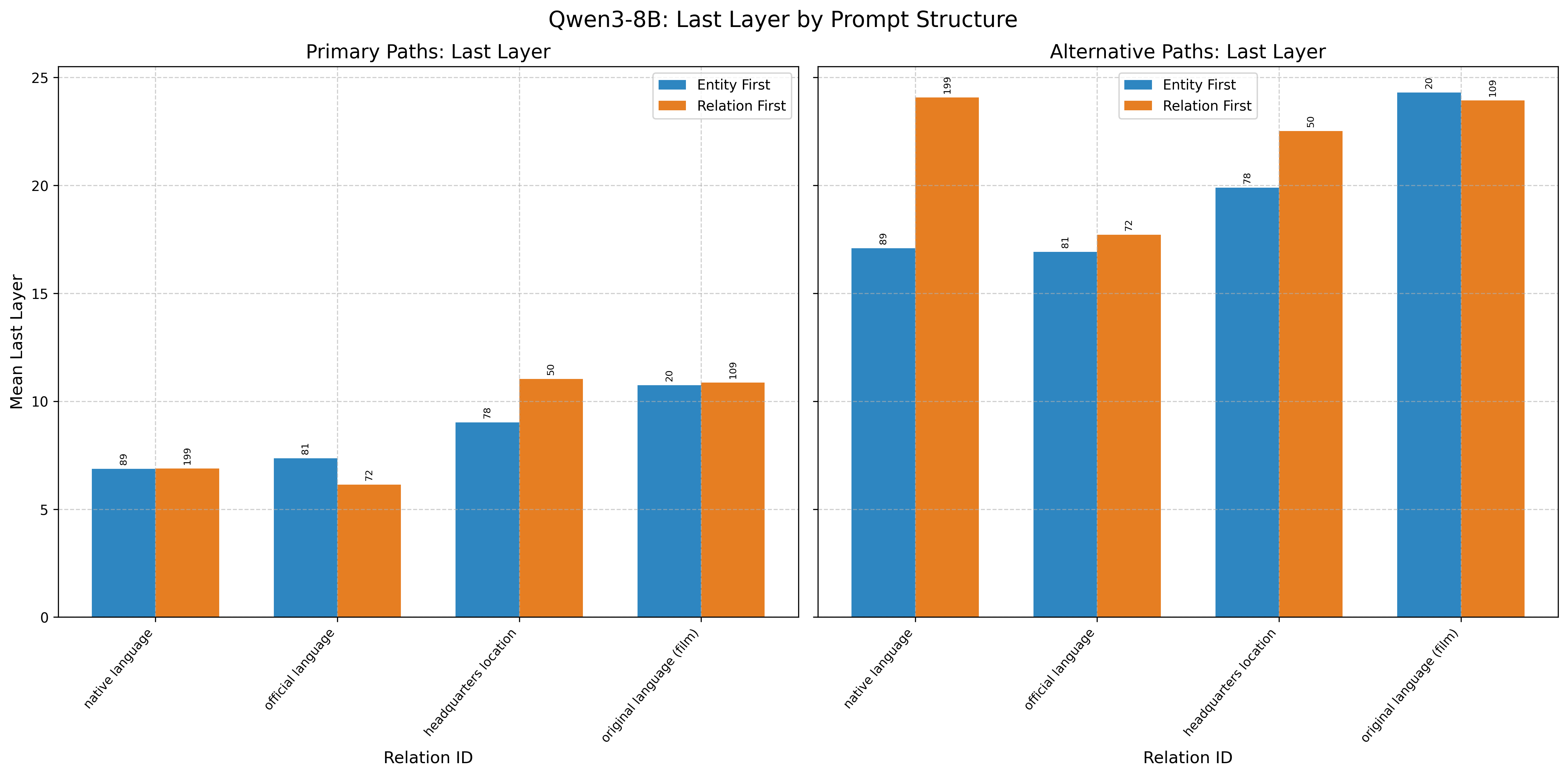}
    }

    \caption{
        Analysis of $\ell_{attr}$ and Path Lengths by Prompt Structure.
    }
    \label{fig:prompt_struct}
\end{figure*}

\begin{figure*}[t!]
    \ContinuedFloat
    \centering
    \subfigure[Path Length vs. Entity Position - Qwen3 8B.]{
        \includegraphics[width=0.8\textwidth]{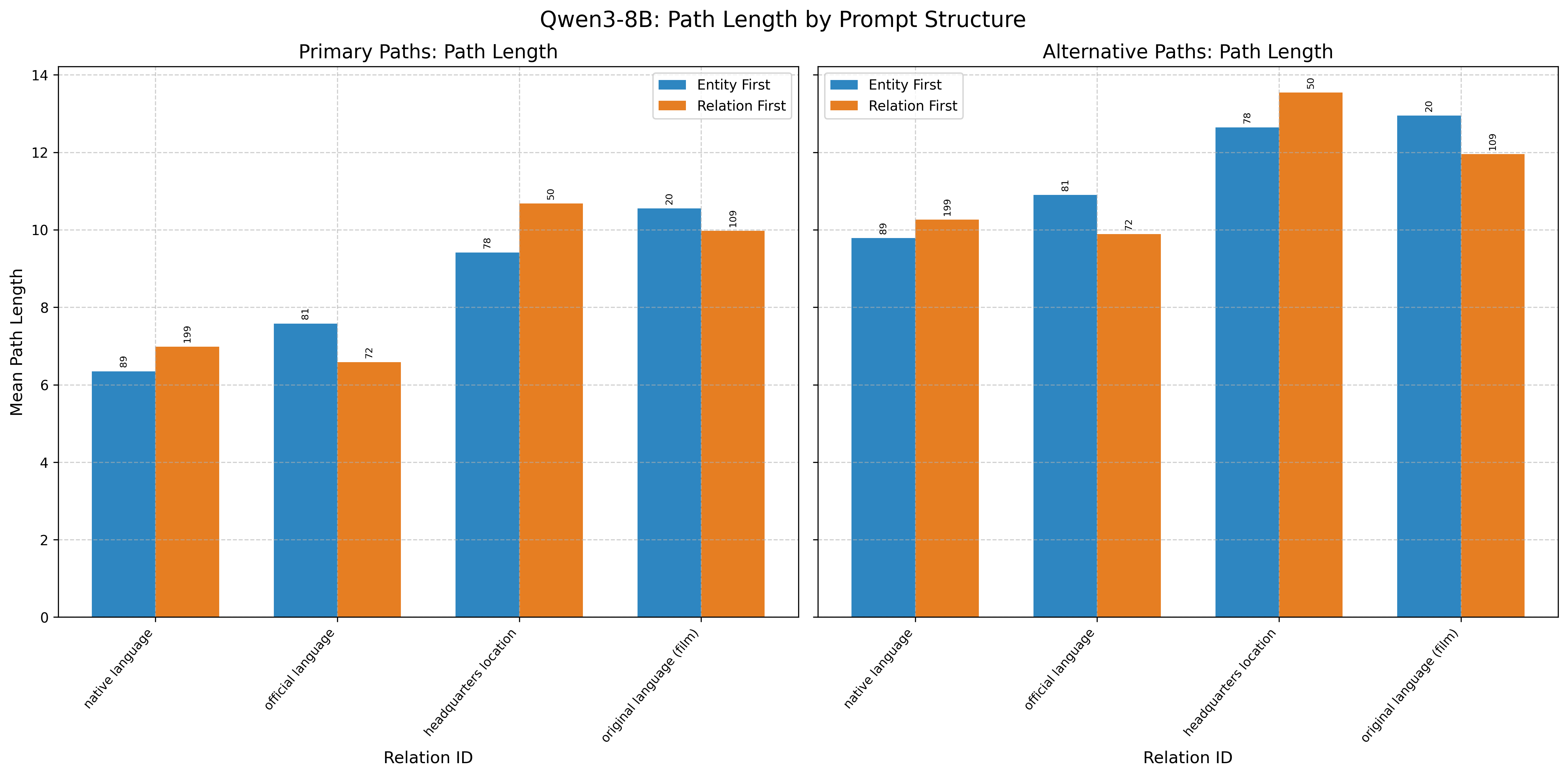}
    }
    \caption{Analysis of $\ell_{attr}$ and Path Lengths by Prompt Structure. (continued)}
\end{figure*}

\section{Relation-Specific Path Analysis}
\label{app:rel_analysis}

\begin{figure*}[t]
    \centering
    \includegraphics[width=\textwidth]{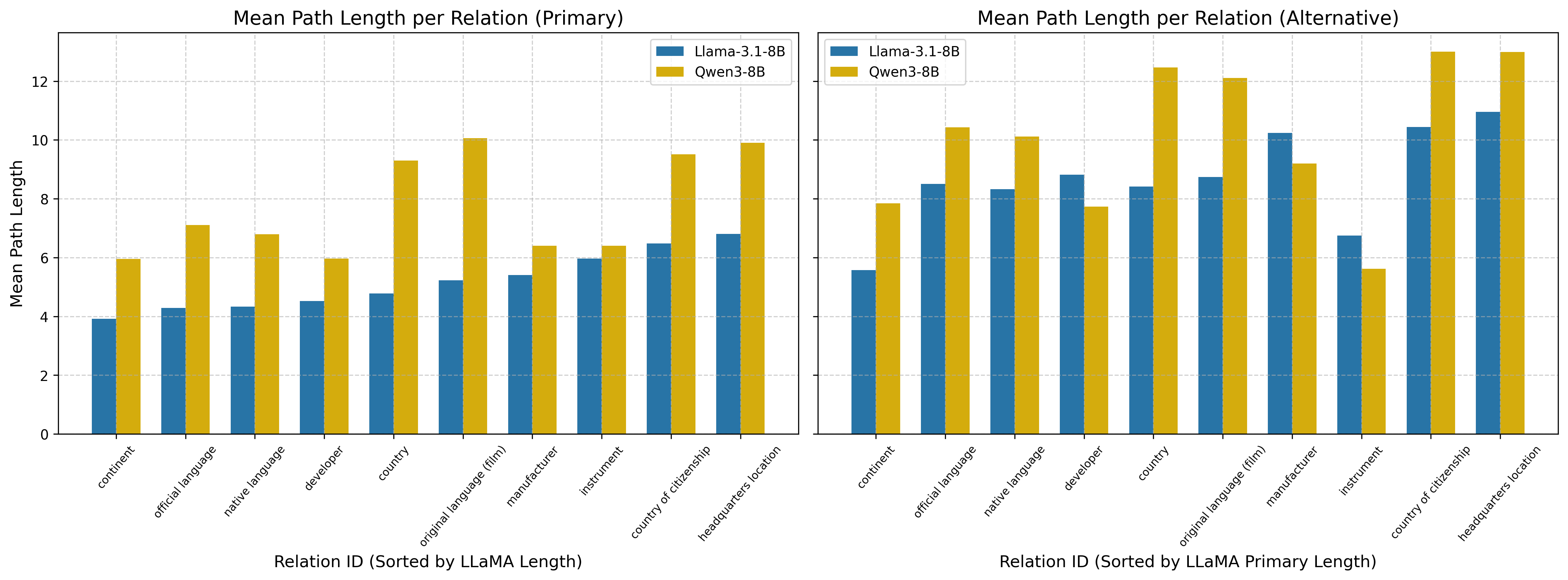}
    \caption{Mean path length per relation. Relations are sorted by LLaMA primary path length.}
    \label{fig:mean_path_length}
\end{figure*}

\begin{figure*}[h]
    \centering
    \includegraphics[width=0.95\textwidth]{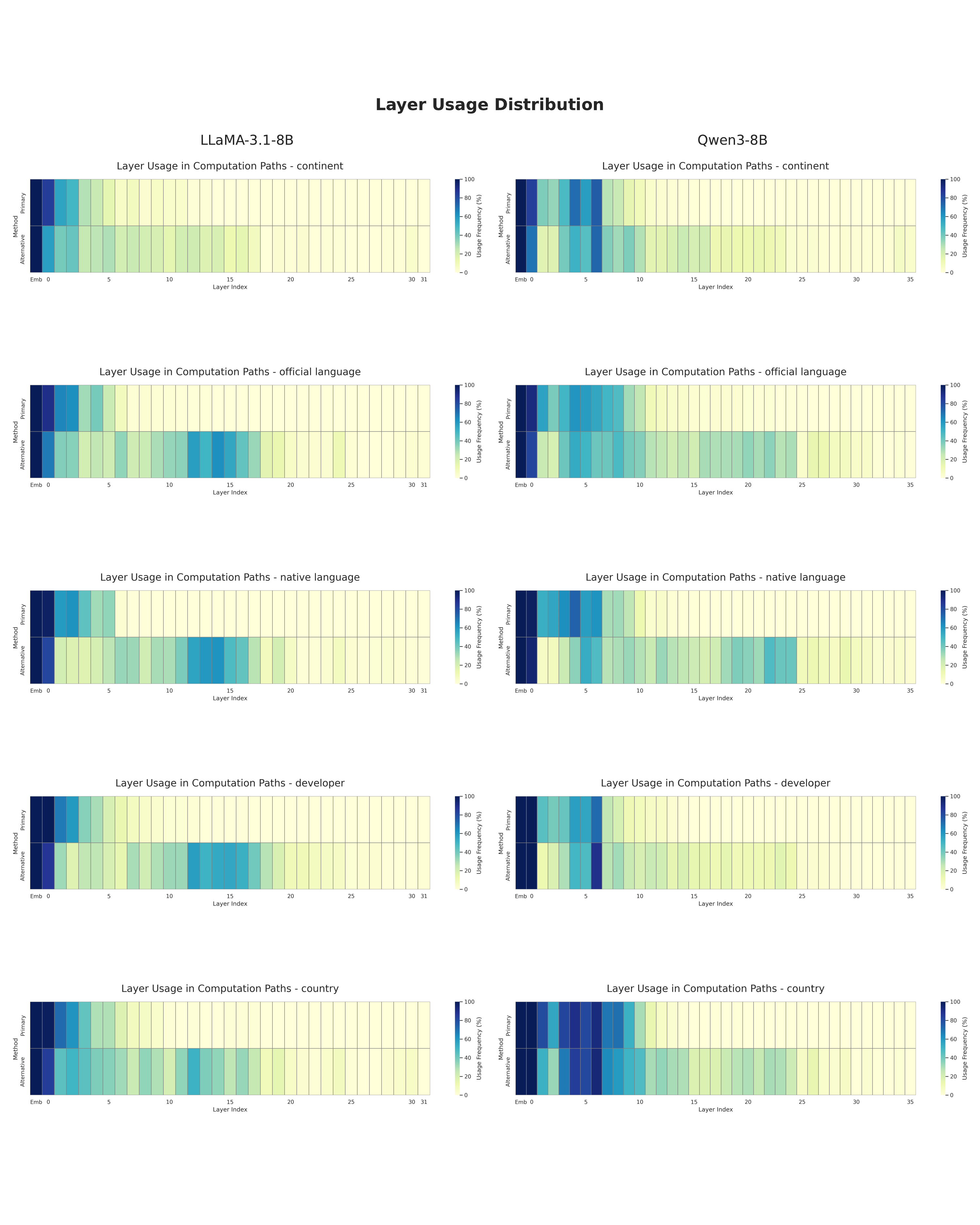} 
    \caption{\textbf{Layer Usage per Relation (Part 1).} Comparison of Relations 1–5. LLaMA 3.1 8B (Left) vs. Qwen3 8B (Right). Warmer colors indicate higher usage probability.}
    \label{fig:heatmaps_part1}
\end{figure*}

\begin{figure*}[h]
    \centering
    \includegraphics[width=0.95\textwidth]{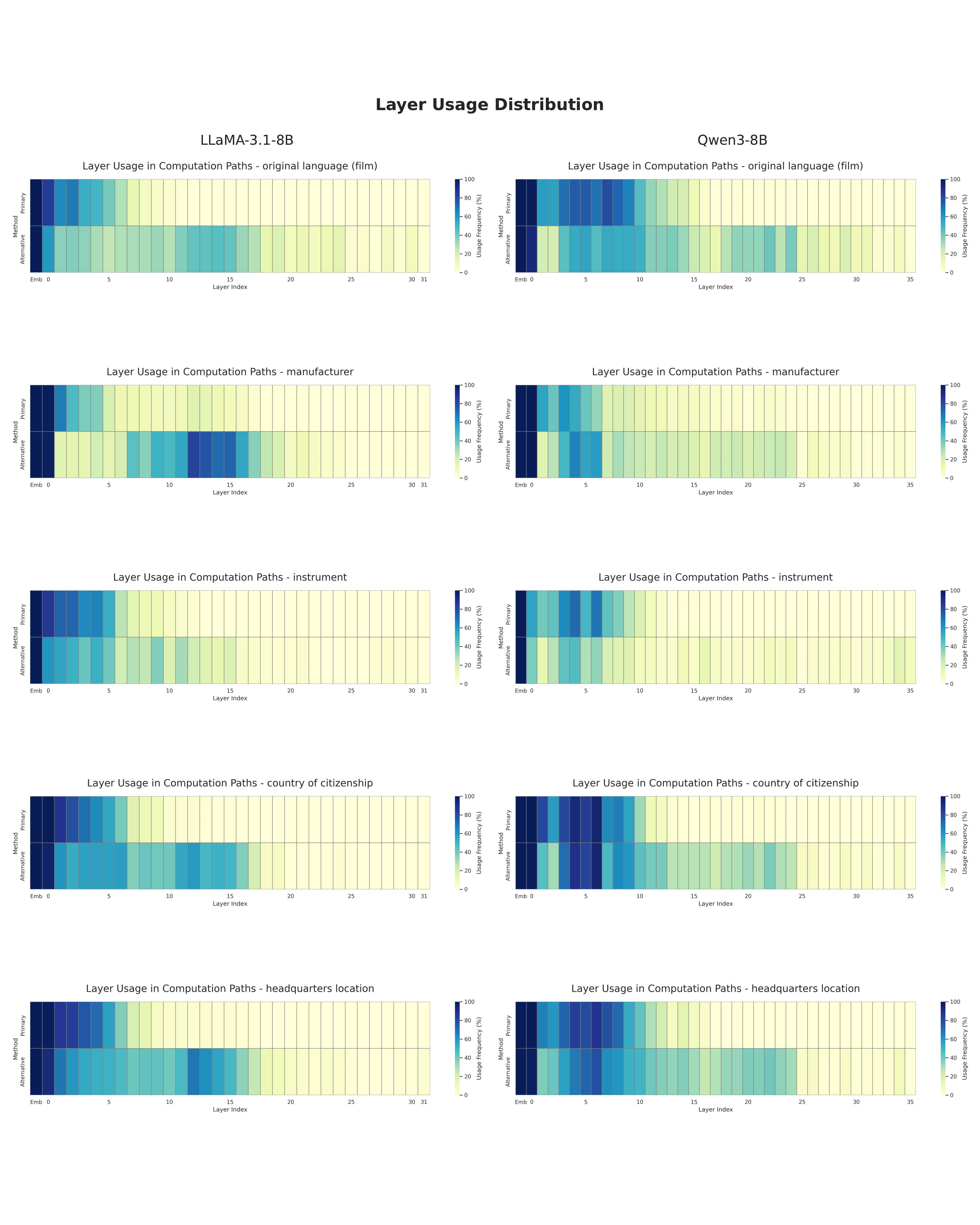}
    \caption{\textbf{Layer Usage per Relation (Part 2).} Comparison of Relations 6–10.}
    \label{fig:heatmaps_part2}
\end{figure*}

In Section \ref{subsec:additional_exp}, we discussed how path length varies by relation. This section provides a comprehensive breakdown of these variations. First, Figure \ref{fig:mean_path_length} illustrates the mean path length across all relations, sorted by the primary path length in LLaMA 3.1 8B. Following this overview, Figures \ref{fig:heatmaps_part1} and \ref{fig:heatmaps_part2} present the granular layer-usage distributions for these relations, comparing LLaMA 3.1 8B and Qwen3 8B.

\section{Entity Resolution Detection Results}
\label{app:entity_resolution}

In Section \ref{subsec:additional_exp}, we analyzed the relationship between the attribute computation path and the explicit resolution of the entity's identity. 

To perform this analysis, we employed the Patchscopes method using the identical constant prompt configuration from \citealp{ghandeharioun2024patchscopes}. The specific few-shot prompt is:
\begin{quote}
    ``Syria: Syria is a country in the Middle East, Leonardo DiCaprio: Leonardo DiCaprio is an American actor, Samsung: Samsung is a South Korean multinational corporation, $x$''
\end{quote}
where $x$ serves as the placeholder for the injected representation.

\begin{figure}[ht]
    \centering
    \includegraphics[width=0.95\linewidth]{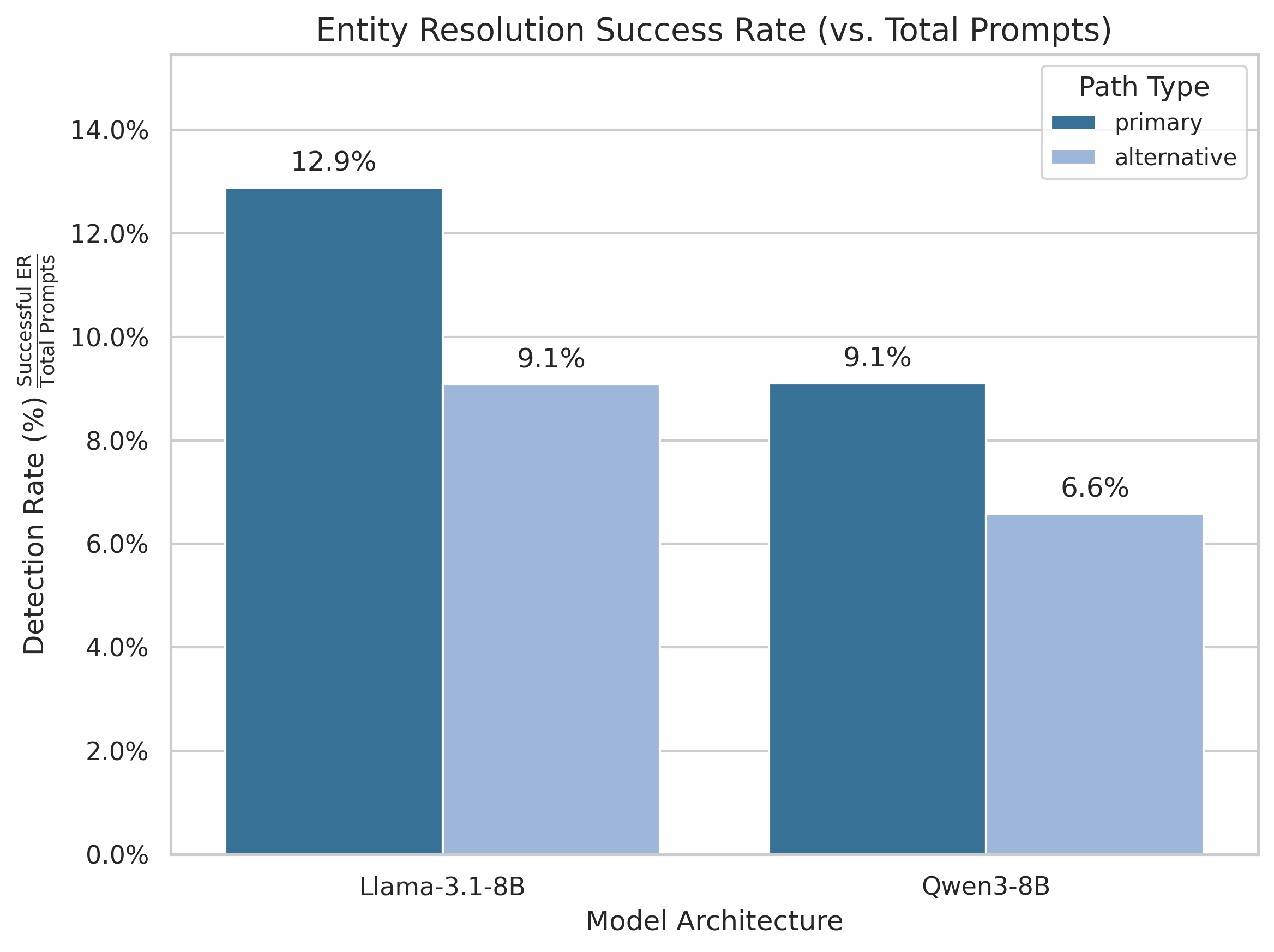}
    \caption{\textbf{Entity Resolution (ER) Detection Success Rate.}
    The figure shows the percentage of prompts where entity resolution was successfully detected along the minimal computation path (relative to the prompts analyzed for each method and model).}
    \label{fig:ent_res}
\end{figure}

Figure \ref{fig:ent_res} presents the quantitative results of this analysis. It details the success rate of detecting Entity Resolution (ER) along both the primary and alternative computation paths for LLaMA 3.1 8B and Qwen3 8B.

\section{Impact of Counterfactual Noise}
\label{app:noise_free_heatmaps}
In this section, we provide the detailed layer-wise usage heatmaps for both LLaMA 3.1 8B and Qwen3 8B under the modified \emph{isolate} operation, where excluded layers are supplied with the output of the most recent preceding path layer rather than counterfactual noise.

\begin{figure}[ht]
    \centering

    \subfigure[LLaMA 3.1 8B]{
        \includegraphics[width=1.0\columnwidth]{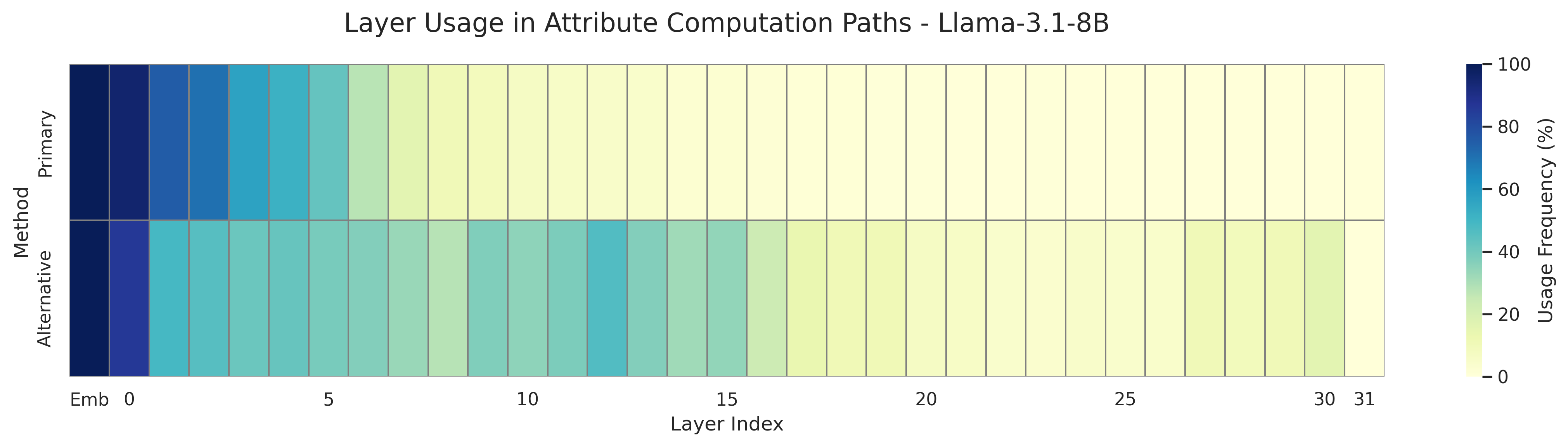}
    }

    \vspace{0.3cm}

    \subfigure[Qwen3 8B]{
        \includegraphics[width=1.0\columnwidth]{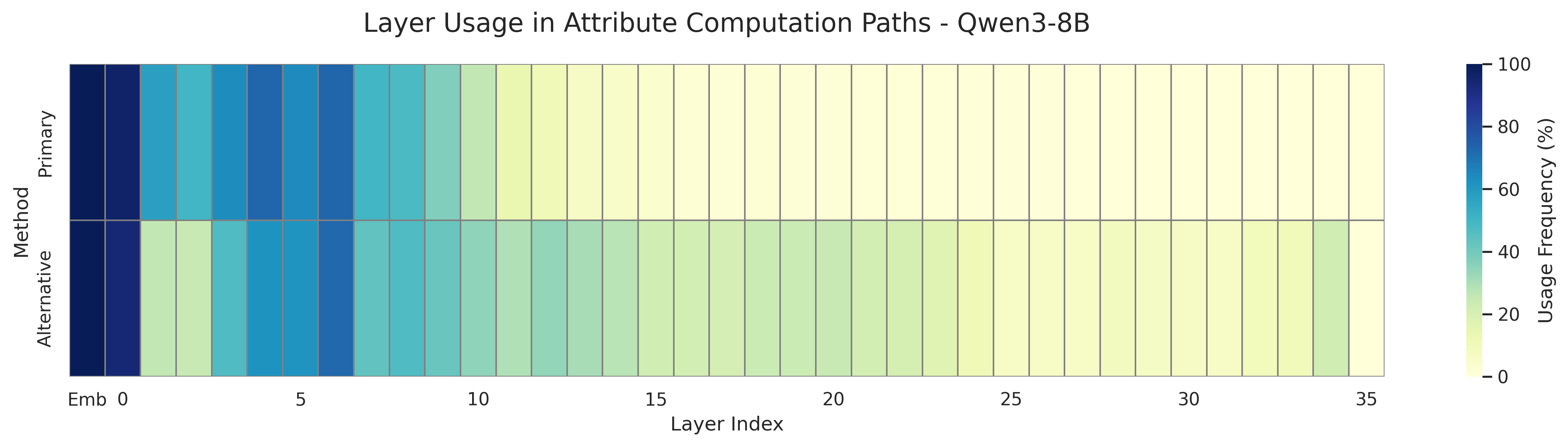}
    }

    \caption{Layer usage heatmaps for primary and alternative paths in LLaMA 3.1 8B and Qwen3 8B under the noise-free setting.}
    \label{fig:noise_free_heatmaps}
\end{figure}

The heatmaps in Figure \ref{fig:noise_free_heatmaps} illustrate the frequency (as a percentage) at which each specific layer is incorporated into the identified primary and alternative computation paths across the dataset.

Consistent with the findings discussed in Section \ref{subsec:additional_exp}, these visualizations highlight a distinct structural shift in the alternative paths when counterfactual noise is removed. While primary path layer distributions remain relatively stable, the alternative paths in both models exhibit a distinct increase in the utilization of deeper layers. Specifically, LLaMA demonstrates a concentrated increase in utilization across layers 27–30, while Qwen exhibits a sharp spike at 22\% in its penultimate layer (layer 34). These visual patterns corroborate the hypothesis that without the forced intervention of counterfactual noise, our greedy search naturally defers the attribute computation to much deeper layers.

\end{document}